\documentclass[12pt]{article}
\usepackage[dvips]{graphicx}
\usepackage{ulem}
\usepackage{setspace}
\usepackage{multirow}
\usepackage{amssymb}
\usepackage{amsmath,subfigure}
\usepackage[small]{caption}
\title{Fast SVM-based Feature Elimination Utilizing Data Radius, Hard-Margin, Soft-Margin}
\author{Yaman Aksu}
\def \bzero {{\bf{0}}}
\def \bone {{\bf{1}}}
\def \bB {{\bf{B}}}

\def \bG {{\bf{G}}}
\def \bM {{\bf{M}}}

\def \bV {{\bf{V}}}

\def \bd {{\bf{d}}}

\def \be {{\bf{e}}}

\def \bs1 {{\bf{s_1}}}
\def \bsk {{\bf{s_k}}}
\def \bsl {{\bf{s_l}}}

\def \bt {{\bf{t}}}

\def \bu {{\bf{u}}}
\def \bui {{\bf{u_{i}}}}
\def \bx {{\bf{x}}}
\def \bxi {{\bf{x_i}}}
\def \bxj {{\bf{x_j}}}
\def \bxn {{\bf{x_n}}}

\def \by {{\bf{y}}}
\def \bXi {{\underline{\xi}}}
\def \bw {{\bf{w}}}

\def \ldotswithhspace {\hspace{0.05in} \ldots \hspace{0.05in}}

\def \uxi {{\underline{\xi}}}
\def \uphi {{\underline{\phi}}}

\def \bzero {{\bf{0}}}
\def \bone {{\bf{1}}}

\def \bA {{\bf{A}}}
\def \bAI {{\bf{A^{\Ical}}}}

\def \bAW {{\bf{A^W}}}
\def \bAWzero {{\bf{A^{W^{0}}}}}
\def \bAWk {{\bf{A^{W^{k}}}}}

\def \bAWoneone {{\bf{A^{W_{11}}}}}
\def \bAWonetwo {{\bf{A^{W_{12}}}}}
\def \bAWtwotwo {{\bf{A^{W_{22}+N}}}}
\def \bB {{\bf{B}}}

\def \bG {{\bf{G}}}
\def \bI {{\bf{I}}}

\def \bIoneone {{\bf{I^{W_{11}}}}}
\def \bIonetwo {{\bf{I^{W_{12}}}}}
\def \bItwotwo {{\bf{I^{W_{22}}}}}
\def \bM {{\bf{M}}}

\def \bV {{\bf{V}}}

\def \bVoneone {{\bf{V^{W_{11}}}}}
\def \bVonetwo {{\bf{V^{W_{12}}}}}
\def \bVonetwoind {{\bf{V^{W_{12}}_{ind}}}}
\def \bVonetwodep {{\bf{V^{W_{12}}_{dep}}}}

\def \bZ {{\bf{Z}}}
\def \bZk {{\bf{Z^k}}}
\def \bZtilde {{\bf{\tilde{Z}}}}
\def \bai {{\bf{a_{i}}}}

\def \bd {{\bf{d}}}

\def \be {{\bf{e}}}

\def \bhk {{\bf{h_{k}}}}

\def \bn {{\bf{n}}}
\def \bnZ {{\bf{n_{z}}}}

\def \bpk {{\bf{p_{k}}}}
\def \bpkstar {{\bf{p_{k}^{*}}}}
\def \bpkin {{\bf{p_{k}^{in}}}}
\def \bq {{\bf{q}}}
\def \bqzero {{\bf{q_{0}}}}

\def \bqonetwo {{\bf{q^{ot}}}}

\def \bqstar {{\bf{q^{*}}}}

\def \bqk {{\bf{q_{k}}}}

\def \bqkplusone {{\bf{q_{k+1}}}}
\def \brk {{\bf{r_k}}}
\def \bs1 {{\bf{s_1}}}
\def \bsk {{\bf{s_k}}}
\def \bsl {{\bf{s_l}}}

\def \bt {{\bf{t}}}

\def \bu {{\bf{u}}}

\def \bvone {{\bf{v_{1}}}}

\def \bvn {{\bf{v_n}}}

\def \bvN {{\bf{v_{N}}}}
\def \bx {{\bf{x}}}

\def \bxn {{\bf{x_n}}}

\def \by {{\bf{y}}}
\def \bXi {{\underline{\xi}}}
\def \bw {{\bf{w}}}

\def \bzrowone {{\bf{z_{r1}}}}
\def \ldotswithhspace {\hspace{0.05in} \ldots \hspace{0.05in}}

\def \uxi {{\underline{\xi}}}
\def \uphi {{\underline{\phi}}}

\newcommand{\Acal}{\mathcal{A}}
\newcommand{\Ccal}{\mathcal{C}}

\newcommand{\Ical}{\mathcal{I}}

\newcommand{\Mcal}{\mathcal{M}}
\newcommand{\Ncal}{\mathcal{N}}
\newcommand{\Rcal}{\mathcal{R}}
\newcommand{\Scal}{\mathcal{S}}

\newcommand{\AWblocks}{ \left(
  \begin{array}{cc}
    \bAWoneone \\
    \bB \\
    \bAWtwotwo \\
  \end{array}
\right) }
\newcommand{\Bblockspostsub}{ \left(
  \begin{array}{cc}
    \bVonetwo \hspace{0.2in} \bzero \\
    \bzero \hspace{0.3in} \bIonetwo \\
  \end{array}
\right) }
\begin{document}
\small
\maketitle
\begin{abstract}
Margin maximization in the hard-margin sense, proposed as feature
elimination criterion by the MFE-LO method, is combined here with
data radius utilization to further aim to lower generalization
error, as several published bounds and bound-related formulations
pertaining to lowering misclassification risk (or error) pertain to
radius e.g. product of squared radius and weight vector squared
norm. Additionally, we propose additional novel feature elimination
criteria that, while instead being in the soft-margin sense, too can
utilize data radius, utilizing previously published bound-related
formulations for approaching radius for the soft-margin sense,
whereby e.g. a focus was on the principle stated therein as
``finding a bound whose minima are in a region with small
leave-one-out values may be more important than its tightness''.
These additional criteria we propose combine radius utilization with
a novel and computationally low-cost soft-margin light classifier
retraining approach we devise named QP1; QP1 is the soft-margin
alternative to the hard-margin LO. We correct an error in the MFE-LO
description, find MFE-LO achieves the highest generalization
accuracy among the previously published margin-based feature
elimination (MFE) methods, discuss some limitations of MFE-LO, and
find our novel methods herein outperform MFE-LO, attain lower test
set classification error rate. On several datasets that each both
have a large number of features and fall into the `large features
few samples' dataset category, and on datasets with lower
(low-to-intermediate) number of features, our novel methods give
promising results. Especially, among our methods the tunable ones,
that do not employ (the non-tunable) LO approach, can be tuned more
aggressively in the future than herein, to aim to demonstrate for
them even higher performance than herein.
\end{abstract}
\vspace{-0.1in}
\section{Introduction} \label{sec:intro}
For information on support vector machines (SVMs), interested
readers can be referred to e.g. \cite{Burges_SVMtut},
\cite{Hastie_book}, \cite{Aksu_TNN}. Our brief summary of SVMs below
gives notation for our manuscript, which is similar to the notation
in \cite{Aksu_TNN}.

The labeled training data is $\{(\bxn,y_n), n\in\Ncal\}$ where
$\Ncal\equiv\{1, \ldots ,N\}$; sample $\bxn$
$\in \mathbb{R}^M$ has class label $y_n\in\{\pm 1\}$. $f(\bx) \equiv
\bw^{{\rm T}} \bx + w_0$, $\bw\in \mathbb{R}^M$, $w_0\in
\mathbb{R}$, is a hyperplane acting as a two-class decision
function. With $g_n\equiv g(\bxn) \equiv y_n f(\bxn)$,
$\frac{g_n}{||\bw||}$ is the signed distance from $\bxn$ to the
decision boundary which is a separating one if $g_n
> 0 \hspace{0.02in}\forall n$ with margin defined
as $\gamma \equiv \frac{\min_{n} g_n}{||\bw||}$. Hard-margin SVM is
a linear or generalized linear two-class classifier defined via the
optimization problem
\begin{equation}
\label{eqn:handle-sep} \min_{\bw,w_0} \frac{1}{2} ||\bw||^2
\hspace{0.03in} s.t. \hspace{0.03in} y_n f(\bxn) \geq 1, \forall n
\end{equation} and soft-margin SVM
is a linear or generalized linear two-class classifier defined via
the optimization problem
\begin{equation}
\label{eqn:handle-nonsep} \min_{\bw,w_0,\uxi} \frac{1}{2} ||\bw||^2
+ C\sum\limits_{n=1}^{N}\xi_n \hspace{0.03in} s.t. \hspace{0.03in}
\xi_n \geq 0, \hspace{0.03in} y_n f(\bxn) \geq 1-\xi_n, \forall n
\end{equation}
In the linear case, the SVM weight vector is given by $\bw \equiv
\sum\limits_{k\in \Scal} \lambda_{\bsk}y_{\bsk} \bsk$, where
$S=\{\bsk: \bf{k}\in\Scal\equiv\{1,...,T\}$), used to specify the
SVM solution, is the set of support vectors which is a subset of the
training points, and $\lambda_{\bsk}$ are the associated
Lagrange multipliers.

The generalized linear (nonlinear) case involves nonlinear functions
$\phi_i(\cdot)$ and $\uphi(\bx) \equiv [\phi_1(\bx),
\ldots,\phi_L(\bx)]^{{\rm T}}$. Inner products between $\uphi(\bx)$
and $\uphi(\bu)$ that can be efficiently computed via a positive
definite kernel function $K(\bx,\bu) \equiv \uphi^{\rm
T}(\bx)\uphi(\bu)$ are of particular interest; in this case,
$\uphi(\cdot)$ and $\bw$ need not be explicitly defined since both
the SVM discriminant function $f$ and the weight vector
squared 2-norm can be expressed solely in terms of the kernel:
\begin{equation}
\label{eqn:fnonlin} f(\bx) = \sum\limits_{k\in \Scal}
\lambda_{\bsk}y_{\bsk} K(\bsk,\bx) + w_0
\end{equation}
\begin{equation}
\label{eqn:wnormsquared} ||\bw||^2 = \sum\limits_{k\in
\Scal}\sum\limits_{l\in \Scal} \lambda_{\bsk}y_{\bsk}
\lambda_{\bsl}y_{\bsl}K(\bsk,\bsl).
\end{equation}

This ``kernel trick'', where a specified $K$ is provided to SVM
training, is the nonlinear kernel case.

Relating these SVM concepts to feature elimination algorithms,
\cite{Aksu_TNN} proposed a so-called ``strict margin maximization''
(or, margin maximization in the hard-margin sense) method called
`basic MFE' that picks the feature elimination that preserves
maximum (positive) margin in the reduced space as
follows\footnote{Notation: During feature elimination, one or more
features can be eliminated in one `step'; i.e. the elimination is
`stepwise'. $\Rcal$ denotes the retained feature set at the start of
a step. We denote a quantity (or variable) $q$ under the step's
(candidate or actual) elimination of a set $\Mcal$ of features in
multiple equivalent ways: $q^{-\Mcal}$ (i.e.
$q^{\Rcal\backslash{\Mcal}}$) to simply only convey $\Mcal$;
$q^{i,-\Mcal}$ to convey the step index $i$, at the left of $\Mcal$;
$q^{-\Mcal,n_a}$ to convey that the sample with the sample index
$n_a$ is being considered a `margin-setter' (aka `anchor') sample
(discussed below), at the right of $\Mcal$. Use of superscript $-m$
as an alternative to $-\Mcal$, where $m$ is feature index (for a
single feature), refers to 1-by-1 elimination of features.}: $m^* =
\displaystyle\mbox{arg}\max_{m \in \{\tilde{m}\in
\Rcal|g_l^{\tilde{m}}>0 \forall l\}} \min_{n}
g_n^{-m}/||\bw^{-m}||$; and a second, counterpart method (MFE-Slack)
based on generalization of strict margin maximization that picks the
feature elimination with the smallest SVM objective function
(\ref{eqn:handle-nonsep}) in the reduced space via the discrete
optimization problem $(m^*, n^*) = \displaystyle\mbox{arg}\min_{m
\in \Rcal } \min_{n_a\in\{l|g_{l}^{-m}>0\}} \frac{1}{2}
(||\bw||^2)^{-m} (\rho^{-m,n_a})^2 +
C\sum\limits_{n=1}^{N}\xi_{n}^{-m,n_a}$; for each candidate $m$ for
elimination, every (correctly classified) sample is evaluated as the
potential margin-setter $n_a$, with both the weight vector squared
norm (WVSN) and slacknesses evaluated  post-feature-elimination, to
pick the optimal $(m^*, n^*)$ that, post-elimination, minimizes
(\ref{eqn:handle-nonsep}) over all discrete choices $\{(m,n_a)\}$. A
hybrid \cite{Aksu_TNN} (here MFEh) used `basic MFE' when applicable
(at steps data is separable) and MFE-Slack at other steps.

\section{Related work: Little Optimization (LO)}
To increase with little computation the {\it margin maximization in
the hard-margin sense} that `basic MFE' can obtain alone (in reduced
space, under weights $(\bw, w_{0})$), the LO approach
\cite{Aksu_TNN} considered the parameterization $(a \bw, b)$ where
$a$ and $b$ are scalars to be optimized, with $\bw$ held fixed;
\textit{i.e.} posed the standard hard-margin SVM training problem
but optimizing in this two-d $(a,b)$ space:
\begin{equation} \label{eqn:lo-lin} \min_{a,b} a^2
s.t. \hspace{0.03in} y_n (a({\bw}^{\rm T}\bxn) + b) \geq 1, \forall
n; \end{equation}
\underline{E}mbedding LO into the elimination
decision (to eliminate by largest post-LO margin in reduced space)
\cite{Aksu_TNN} is herein referred to as MFE-LOe (or MFE-LO in some
graphs).

\textbf{Correction to LO:} Before continuing, we now point out an
error in how \cite{Aksu_TNN} solves (\ref{eqn:lo-lin}) and correct
the error. To give an intuitive graphical description, we now refer
to Fig. 4 in \cite{Aksu_TNN}. In the illustrated halfspace $a>0$,
the entirety of the illustrated shaded region (defined by (i.e.
lying to the right of the intersection point of) the two thick
lines) is not the correct feasible region of the problem; the
correct feasible region is the smaller (shaded) cone defined by
(i.e. lying to the right of the intersection point of) the two thin
lines (one solid (going through $w_0=1$), one dashed (going through
$w_0=-1$)). Notice that accordingly the statement in \cite{Aksu_TNN}
that the feasible region is defined by the cone ``bounded by the
line $l_{2}^+$ with maximum slope in $L_2$ and the line $l_{1}^+$
with minimum slope in $L_1$'' is incorrect. The (correct) feasible
region is defined by the cone bounded by the line with {\it minimum}
slope in $L_2$ and the line with {\it maximum} slope in $L_1$. The
LO solution, i.e. the (feasible) minimum $a^2$, lies at that cone's
tip; this tip is shown in the Figure as the intersection point of
two lines immediately above the ``$\Ccal$+'' label shown in the
Figure.

\begin{figure*}
\centering \subfigure[] { \label{fig:test-sel1_mfelo-vs-mfeh}
   \includegraphics[scale=.5]{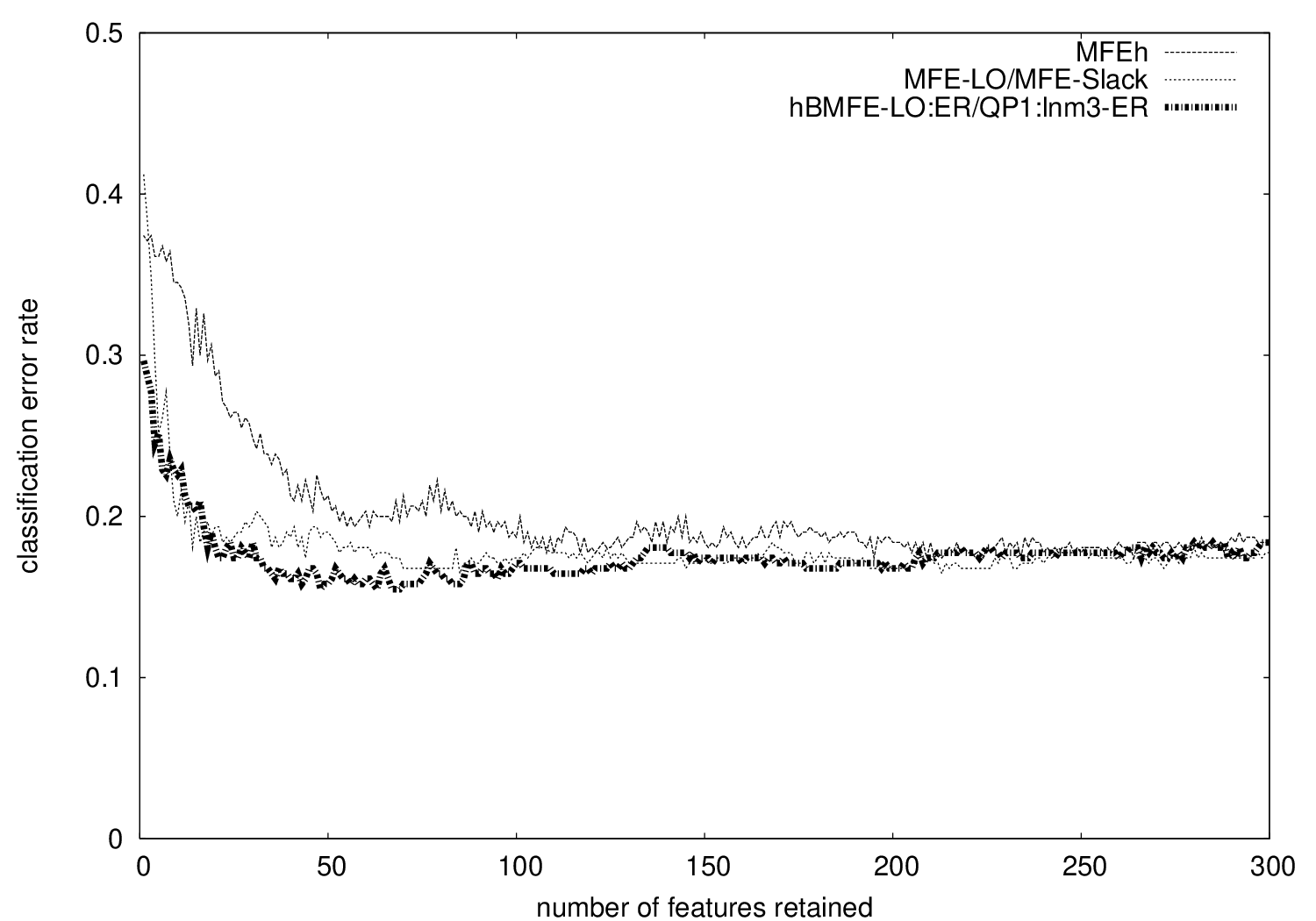}
} \subfigure[] { \label{fig:train-sel1_mfelo-vs-mfeh}
   \includegraphics[scale=.5]{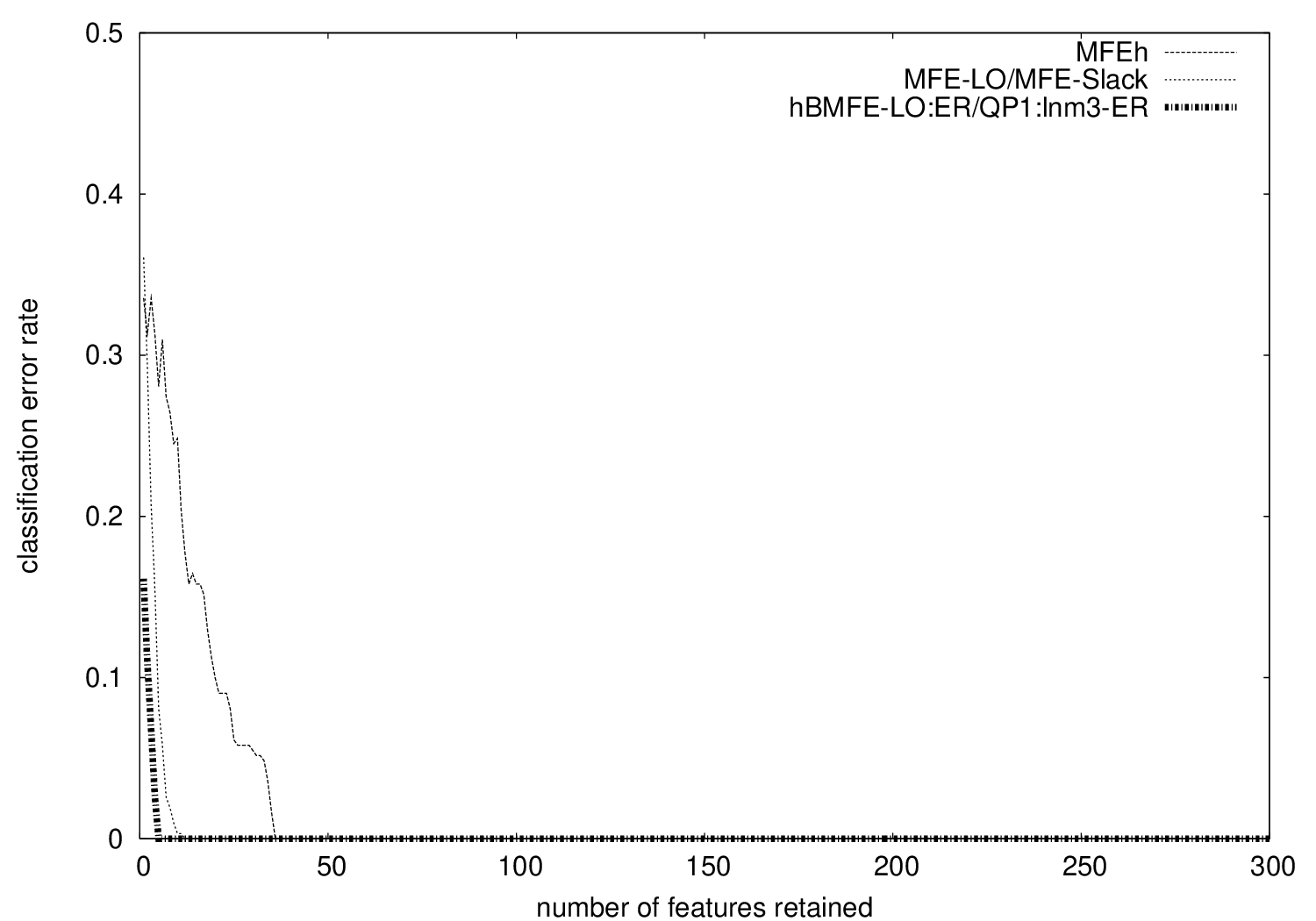}
} \caption[stuff]{(a) MFE-LOe achieved much lower across-trial
average {\it test} set classification error rate i.e. better
generalization than MFEh. (b) The (across-trial average) {\it
training} set classification error rate curves illustrate that
initially separable data remained separable longer under MFE-LOe
than MFEh. SVM linear kernel case; Colon Cancer gene dataset with
2000 features and much fewer samples. Here, for the particular case
of 1-by-1 elimination of features.} \label{fig:sel1_mfelo-vs-mfeh}
\end{figure*}
\begin{figure*}
\centering \subfigure[] {
   \includegraphics[scale=.6]{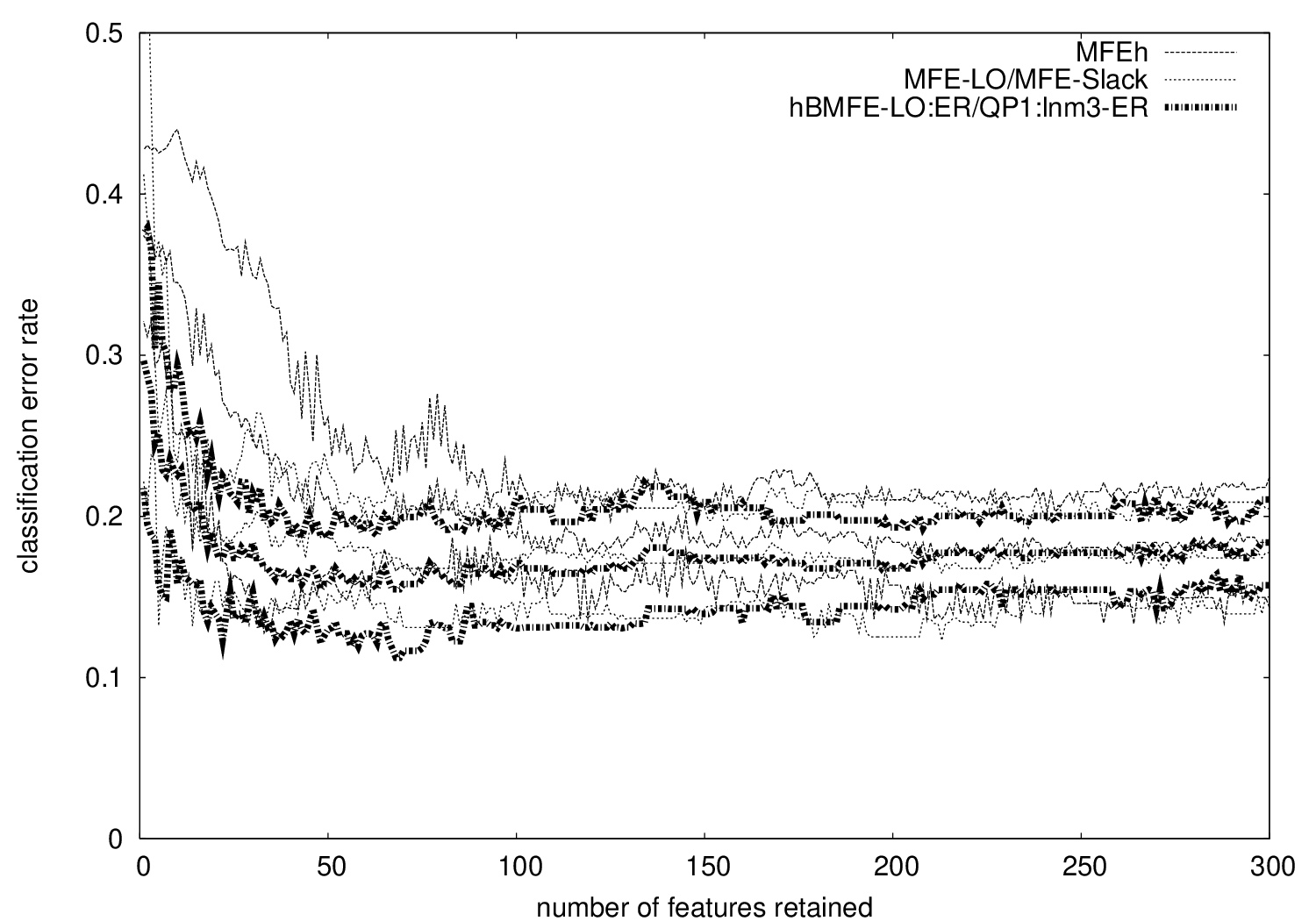}
} \caption[stuff]{} \label{fig:sel1_enhanced_mfelo-vs-mfeh}
\end{figure*}
\begin{figure*}
\centering \subfigure[] {
   \includegraphics[scale=.5]{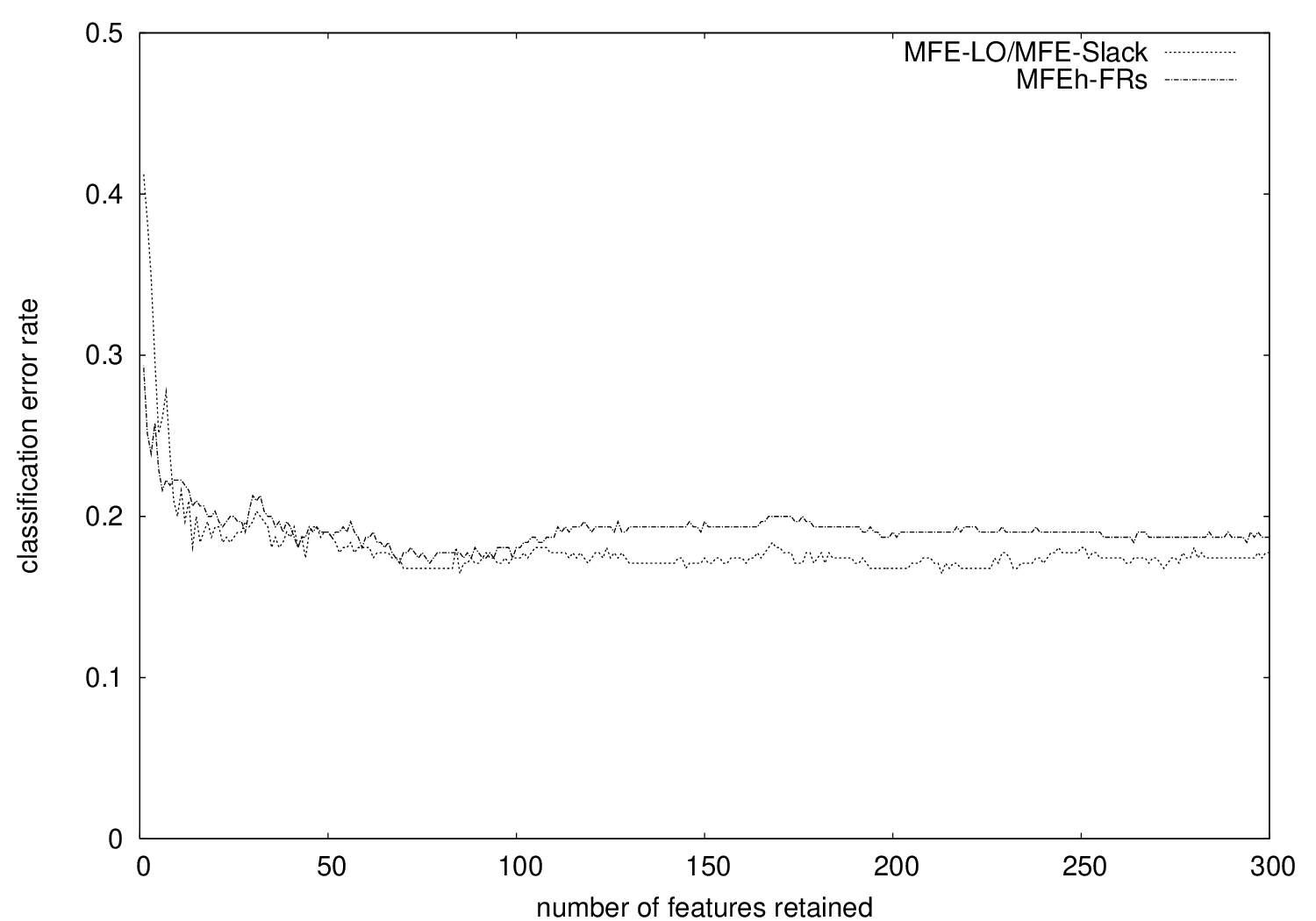}
}  \caption[]{Stepwise light classifier retraining (LOe (MFE-LO))
attaining lower test set classification error rate than stepwise
full SVM retraining (FR (MFEh-FRs)); here, for the particular case
of 1-by-1 elimination of features. SVM linear kernel case; Colon
Cancer gene dataset with 2000 features and much fewer samples.
Zoomed to final 300 features retained.} \label{fig:sel1_lo-vs-fsr}
\end{figure*}
\begin{figure*} \centering \subfigure[] {
   \includegraphics[scale=.4]{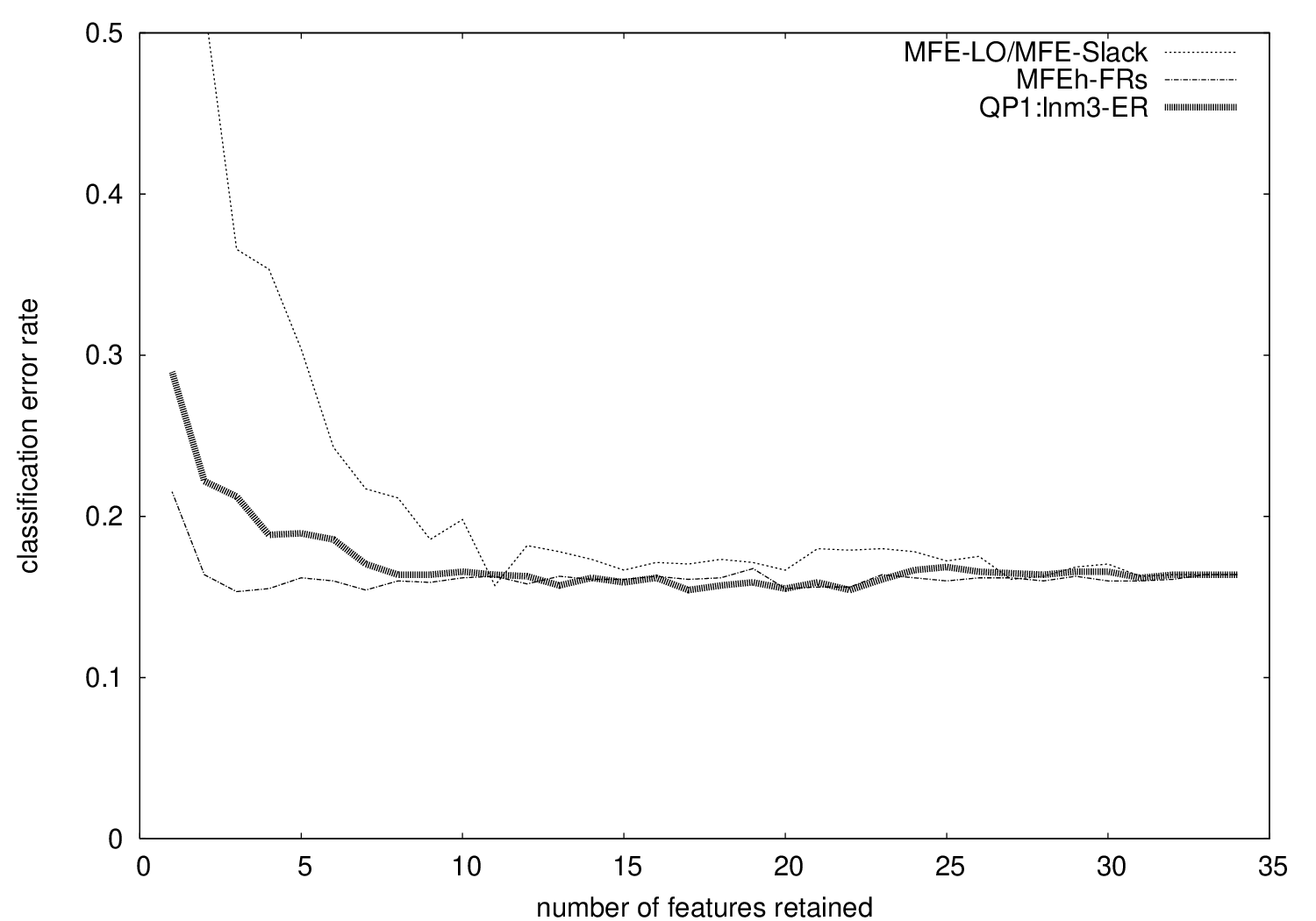}
}  \caption[]{Ionosphere dataset. SVM linear kernel case.}
\label{fig:outperform}
\end{figure*}
LO serves to perform light classifier retraining which has several
generalization accuracy advantages compared to full (SVM) retraining
(FSR aka FR). First, stepwise FSR (to stepwise attain the most
optimal values for SVM margin or objective function) may cause the
subset selection process to overfit; a limitation, especially for a
dataset whose number of features is large since overfitting, a
cumulative effect, is expected when a large number of elimination
steps accumulate. In Fig. \ref{fig:sel1_lo-vs-fsr}, across-trial
average test set classification error rate (see: Sec.
\ref{sec:results} which gives our experiment procedure) is plotted,
as a function of the number of retained features (reduced going from
right to left).\footnote{When in a graph we show two methods paired
with a slash `/', such as MFE-LO and MFE-Slack paired here, the
first and second indicate, respectively, the method used when data
is separable and nonseparable; in this particular Figure, the second
method plays little role within the pairing, since this dataset was
separable until only very few features remained as illustrated by
Fig. \ref{fig:train-sel1_mfelo-vs-mfeh}.} In conjunction with Fig.
\ref{fig:sel1_mfelo-vs-mfeh}, these results for MFEh-FR{\it s}
(stepwise FSR {\underline s}ubsequent to elimination decision by
MFEh) and MFE-LO illustrate that although stepwise FSR can improve
generalization, utilizing FSR throughout a large number of
elimination steps is attaining less generalization accuracy than
light classifier retraining; this result can perhaps be understood
as a type of overfitting. Fig. \ref{fig:sel1_enhanced_mfelo-vs-mfeh}
redraws Fig. \ref{fig:test-sel1_mfelo-vs-mfeh} to supplement each
method's across-trial average curve ($\mu$) with the $\mu+\sigma$
curve (seen above the average curve) and the $\mu-\sigma$ curve
(seen below the average curve), where $\sigma$ is the across-trial
standard deviation; in this manuscript, to demonstrate results more
precisely, we demonstrate standard deviation in this fashion. To
illustrate there may not be much overfitting for a dataset with
hugely fewer features (and hugely lower features-to-samples ratio),
Fig. \ref{fig:outperform} demonstrates FR-based elimination
outperforming elimination based on light classifier retraining
(LO)\footnote{QP1 shown is light classifier retraining that will
shortly be discussed below.}; LO is in both Fig.
\ref{fig:sel1_lo-vs-fsr} (for large number of features) and Fig.
\ref{fig:outperform} (for much smaller number of features), playing
a reference role. Second, FSR can have more computational cost than
light classifier retraining. For initial dimensionality $M$ (e.g.
$7000+$ for gene data), at the {\it i}-th elimination step FSR
trains an SVM for the very large feature dimensionality $M-i$
($6999,6998,\dots$); lower computational cost can be attained by
light classifier retraining that is carried out by LO as well as
carried out by our novel QP1 approach discussed shortly.

We propose and assess several novel feature elimination methods,
including assessing them in comparison with these previously
published MFE methods in \cite{Aksu_TNN} (MFE-LO, MFEh, MFE-Slack),
and discuss the limitations of these previous three methods in doing
so. Since MFE-LO is not usable at any feature elimination step at
which the data is not separable (i.e. a step at which a (pre-LO)
separating classifier is not possible or available), datasets that
remain separable for most (ideally, vast majority) of the steps are
the most suitable datasets for comparing the generalization
performance of MFE-LO to other methods', irrespective of whether or
not these other methods too require separability; especially these
datasets include `small $N$, large $M$' datasets (discussed in
previous work such as \cite{Aksu_TNN}, \cite{Aksu_PLOS},
\cite{Dasgupta_FE}). Accordingly, almost all results given herein,
whenever comparing MFE-LO to our novel methods and to the methods in
\cite{Aksu_TNN}, are for such datasets. In fact, we give results for
each of the three `small $N$, large $M$' gene datasets seen in
\cite{Aksu_TNN} and other works. In Fig.
\ref{fig:sel1_mfelo-vs-mfeh}, across-trial average test set
classification error rate is plotted, as a function of the number of
retained features (reduced going from right to left), illustrating
MFE-LOe outperformed MFEh (and MFE-Slack, whose curve overlaps curve
of MFEh).\footnote{The third method in this Figure will be discussed
in Sec. \ref{sec:radius}.} Extensive results for several datasets
and extensive discussion are given in Sec. \ref{sec:results}.

\section{QP1: slackness-incorporating light classifier retraining} \label{sec:qp1}
We begin by making a central observation, not made in previously
published related work (LO) \cite{Aksu_TNN}: the moment one makes
the modeling assumption that scalars $a$ and $b$ for the
parameterization $(a \bw, b)$ is to be sought while holding $\bw$
fixed (such as made by LO in previous work), what is happening is
that the data to work with is becoming $1$d (scalar); notice in
(\ref{eqn:lo-lin}) that the non-scalar (multi-dimensional) data
variables $\bxn$ (denoted $\bxn^{-\Mcal}$ during the elimination
process) from the original SVM optimization problem are transformed
into scalar data variables ${\bw}^{\rm T}\bxn$
and can thus not only be explored via an optimization formulation
that, unlike LO, is slackness-incorporating but also different ways
to generate a solution for this slackness-incorporating formulation
can be explored. Accordingly, in this Section, we discuss that what
arises from incorporating slackness for such 1d data is a
computationally exceptionally simple quadratic programming (QP)
problem (relatively speaking, considering QP problems in general)
for which a solution can be generated in one of multiple possible
ways including our novel specialized computationally low-cost
active-set method we present (but do not empirically utilize)
herein\footnote{For small- to medium-scale inequality constrained
quadratic programming (ICQP) problems, it has been mentioned that
active-set methods are the most effective \cite{Nocedal} generally;
we took a specific active-set algorithm given in \cite{Nocedal} and
specialized it for our particular novel ICQP problem
(\ref{eqn:qp-linear}) which is discussed shortly, achieving much
computational efficiency for the algorithm in doing so. Herein we
present our active-set method work for mainly as a novel theoretical
mathematical contribution (wherein our devised Lemmas and Theorems
are presented) and do not actually utilize this algorithm in our
current feature elimination experiments herein.}, and we also show
that additional ways to generate a solution conveniently arise from
simply and conveniently employing distinct SVM solver approaches as
we show shortly in this Section that the optimization formulation
(the QP) is equivalent to the simple 1d soft-margin SVM (i.e.
soft-margin SVM for scalar data). For example, when the
abovementioned particular QP (i.e. the QP for scalar data) is
handled as a 1d SVM, one way to generate a solution is to utilize an
SVM solver such as LIBSVM \cite{Libsvm} and another way is
\cite{Su_2002} which too solves the 1d soft-margin SVM problem. This
second way, \cite{Su_2002}, has the built-in limitation that a
support vector (a vector for which the discriminant function value
$yf$ is 1, i.e. a vector ``at the margin''\footnote{As
\cite{Su_2002} stated (see: proof of Observation 4 in
\cite{Su_2002}), ``If $x_i$ is a support vector, then by definition
$y_i(x_i \cdot w+b)=1.$''}) is assigned from within each of the two
classes rather than from within solely one class; this is a
limitation because it narrows the $(a, b, \bXi)$ search space, when
compared with SVM solvers such as LIBSVM and our abovementioned
novel active-set method; in particular, candidate solutions
encountered along the particular descent path that our method takes
as a non-discrete optimization method include candidate solutions
considered by the discrete optimization method \cite{Su_2002}, i.e.
solutions at which $y_i(x_i \cdot w+b)=1$ is simultaneously
fulfilled by {\it two} (training) samples (which, can be thought of
as the ``margin-setter'' samples). To summarize, our mathematical
analytical interest and contributions here are aligned more with the
following three notions collectively than simply aligned with the
more simplistic notion that it is possible to define (and then give
some results for) a slackness version of LO (LO-Slack): 1) making
the abovementioned central ``data to optimize with is now 1d data''
observation, not made by LO, 2) accordingly casting and treating the
problem in a more general setting than LO did, as computationally
low-cost {\underline q}uadratic {\underline p}rogramming for
{\underline 1}d data (QP1), and 3) accordingly providing an analysis
of multiple approaches (each computationally low-cost) that can
generate a solution for this particular setting. Our novel
formulation, (\ref{eqn:qp-linear}) (\ref{eqn:qp-1d}) given below
shortly, is thus quite suitably named QP1, not LO-Slack.\footnote{LO
can be considered a specific type or instantiation of QP1; a special
one that does not incorporate slackness i.e. QP1-NonSlack.}

Given SVM linear weights $(\bw, w_{0})$, we consider the
parameterization $(a \bw, b, \bXi)$, where $a$, $b$, and $N$
slacknesses are scalar parameters to be optimized, with $\bw$ held
fixed. We thus pose the {\it soft-margin} SVM problem
(\ref{eqn:handle-nonsep}) but only optimize in this $(a,b,\bXi)$
parameter space:
\begin{equation} \label{eqn:qp-linear}
\displaystyle\min_{a,b,\bXi} \frac{1}{2} a^2||\bw||^2 +
C\sum\limits_{n}{\xi_n} \hspace{0.03in} s.t. \hspace{0.03in} \xi_{n}
\geq 0, \hspace{0.02in}  y_n ({a\bw}^{\rm T}\bxn + b) \geq 1-\xi_n,
\forall n;
\end{equation}
This formulation, QP1, is distinct from LO wherein, again, it was
only in the hard-margin sense that margin maximization was posed,
motivated, and discussed (focusing on (\ref{eqn:handle-sep}), not
the soft-margin sense (\ref{eqn:handle-nonsep})) i.e. for strictly
satisfying the margin \cite{Aksu_TNN}; while \cite{Aksu_TNN} was
careful to state, by contrast, that ``strictly satisfying the margin
could potentially lead to overfitting when training samples at the
margin are outliers or even mislabeled samples. Optimizing the
amount of slackness (by choosing the parameter $C$), {\it e.g.} via
cross validation, may yield classifiers with better generalization
than those based on strictly maximizing margin.'' The fact that we
formulated QP1 and are analytically discussing multiple
computationally efficient solvers for QP1 herein are a contribution
as it fills a substantial gap left by \cite{Aksu_TNN}.

By contrast to LO, we focus on the soft-margin SVM (\ref{eqn:handle-nonsep});
in our QP1 formulation, since $\bw$ and thus also its norm $||\bw||$ are held
fixed, a change of variables $w\equiv a||\bw||$ and $z_n\equiv{\bw}^{\rm
T}\bxn/||\bw||$ shows the problem is equivalent to the simple 1d
soft-margin SVM (i.e. soft-margin SVM for scalar data) which has
little computational cost:
\begin{equation}
\label{eqn:qp-1d} \displaystyle\min_{w,b,\bXi} \frac{1}{2} w^2 +
C\sum\limits_{n}{\xi_n} \hspace{0.02in} s.t. \hspace{0.02in} \xi_{n}
\geq 0, \hspace{0.01in} y_n (w z_n + b) \geq 1-\xi_n, \forall n
\end{equation}

Since QP1 (\ref{eqn:qp-linear}) (\ref{eqn:qp-1d}) requires little
computation and contains hyperparameters e.g. $C$ as part of its
definition, QP1 can be performed in conjunction with each feature
elimination step and hyperparameter selection can be integrated into
that step. Thus, across, as well as within, elimination steps, one
can generate a set $\{(a,b,\uxi)\}$ of multiple (QP1 output)
triplets i.e. a set $\{(w,b)\}$ of classifiers (herein aka models;
i.e. pairs of scalars $w$ and $b$). However, for the task of picking
among these a particular classifier (with its associated candidate
feature elimination), picking the classifier with the smallest QP1
objective function may not be a great feature elimination criterion;
that particular criterion is not our main focus herein and we
overview it briefly and in an Appendix, so as to now move on to the
notion that QP1 need not form a feature elimination criterion by
itself and can instead, as a type of slackness-incorporating light
classifier retraining (with little computational cost), be combined
with other concepts to define a feature elimination criterion, such
as we do in the upcoming Sec. \ref{sec:radius-soft} where we propose
novel feature elimination criteria that combine QP1 with bounds that
utilize data radius that aim to lower generalization error. A role
of QP1 in such combinations is that QP1 serves to perform light
classifier retraining which has several advantages regarding
generalization accuracy when compared to full SVM retraining (FSR)
as well as when compared to the alternative LO method for light
classifier retraining, as follows. First, stepwise FSR (to stepwise
attain the most optimal values for SVM margin or objective function)
may cause the subset selection process to overfit; a limitation,
especially for a dataset whose number of features is large since
overfitting, a cumulative effect, is expected when a large number of
elimination steps accumulate. To illustrate there may not be much
overfitting for a dataset with hugely fewer features (and hugely
lower features-to-samples ratio), Fig. \ref{fig:outperform}
demonstrates FSR-based methods outperforming light classifier
retraining; see also our earlier above discussion of this Figure.
Second, FSR has more computational complexity than QP1. For initial
dimensionality $M$ (e.g. $7000+$ for gene data), at the {\it i}-th
elimination step, FSR trains an SVM for the large feature
dimensionality $M-i$ ($6999,6998,\dots$), whereas our training
essentially has the computational complexity of a 1d SVM
($1,1,\dots$). Moreover, by incorporating slackness, QP1 does not
require margin maximization in the hard-margin sense whereas LO does
require it; i.e. requiring margin maximization strictly in the
hard-margin sense may lead to overfitting when training samples at
the margin are outliers or even mislabeled samples, as mentioned
above.

Like our QP1 approach, MFE-slack \cite{Aksu_TNN} also incorporates
slackness into the feature elimination criterion. However, it can be
easily noticed that in MFE-slack, unlike in QP1, originally designed
relative magnitudes among SVM Lagrange multipliers and intercept
$w_0$ do not remain unchanged, since MFE-Slack scales a Lagrange
multiplier and $w_0$ by the same scalar (at each feature elimination
step). This is a slight but significant limitation in MFE-slack, as
our Figures demonstrated, which demonstrated that our QP1 approach,
which does modify the abovementioned relative magnitudes (via
jointly optimizing $a$ and $b$ (and the slacknesses $\xi$)), is
outperforming MFE-Slack.

As mentioned above, there are multiple approaches, such as LIBSVM,
\cite{Su_2002}, and our active-set method in the Appendix, that can
generate a solution for the 1d SVM problem (\ref{eqn:qp-1d}); each
is computationally low-cost, including being quite fast. However,
each of these three has its own unique tradeoff between
computational cost and how well the objective function is being
optimized. In our experiments herein, we utilize the first (LIBSVM).

\section{In feature elimination, utilizing bounds that utilize data radius} \label{sec:radius}
In earlier Sections, we discussed an approach that seeks scalars $a$
and $b$ for the parameterization $(a \bw, b)$ while holding $\bw$
fixed (i.e. LO and QP1), whereby the data to work with within a
feature elimination step becomes $1$d (scalar). Next, aiming to
further decrease generalization error, we define novel feature
elimination methods by combining this approach with a utilization of
data radius $R$, essentially the radius of the smallest sphere
containing all $\uphi(\bx)$, because, as we shortly discuss, $R$
appeared as an integral part of several published bounds and
bound-associated optimization formulations for characterizing
generalization error that span the hard-margin and soft-margin
settings. For information on bounds that utilize data radius, we
refer interested readers first to e.g.
\cite{Chung-RadiusMarginBound} and \cite{Weston_NIPS};
\cite{Chung-RadiusMarginBound} focuses on the `radius margin bound'
and `modified radius margin bound' concepts (and associated
optimization formulations) while making useful references to several
other related work on bounds that utilize data radius (e.g.
\cite{Vapnik_1998}, \cite{Vapnik-Chapelle}, \cite{Zhang_radius}),
and \cite{Weston_NIPS} too discusses bounds that utilize data
radius.

\subsection{Utilizing radius in the hard-margin classifier sense} \label{sec:radius-hard}
For the hard-margin classifier case, several published bounds
pertain to the product of squared radius and weight vector squared
norm WVSN ($R^2||\bw||^2$):

1) The first such bound we consider is an upper bound on the VC
dimension $h$. The bound $h < R^2 A^2 +1$ for $h$ (for the function
family $\{f_{\bw,b}: ||\bw|| \leq A\}$ for some scalar $A$) was
discussed in e.g. \cite{Hastie_book,Weston_thesis,Vapnik_1995}.
Lowest upper bound on the VC dimension $h$ is a known criterion for
selecting among multiple functions (for machine learning) a
particular one (so as to aim for lower generalization error), such
as during Structural Risk Minimization (SRM)\footnote{For SRM, see
e.g. \cite{Burges_SVMtut}.}; e.g., when discussing how one can do
SRM, \cite{Burges_SVMtut} asked to find within a set of functions
the particular one that, as \cite{Burges_SVMtut} states, ``gives
maximum margin (and hence the lowest bound on the VC dimension.)''
As seen in feature elimination that seeks to maximize margin in the
hard-margin sense, such as LO, although a WVSN upper bound $A$ (i.e.
$A^{-m}$) is not being computed explicitly, the WVSN itself is
computed, and can be utilized as the available surrogate upper bound
and define a new feature elimination method that picks the feature
elimination for which the product of $R^2$ and WVSN\footnote{The
product $(R^2)^{-m}$WVSN$^{-m}$ in the particular case of 1-by-1
elimination of features; $(R^2)^{-\Mcal}$WVSN$^{-\Mcal}$ generally.}
is smallest\footnote{Of course, by contrast, when not doing feature
elimination and no classifier to start from in order to guide the
elimination is yet available (a scenario that requires all
classifier weights to be simultaneously generated from scratch from
training data), mathematically optimizing the product of squared
radius and WVSN (i.e. the joint optimization of these two
quantities) is not so straightforward (simultaneously with such
generation); e.g. as the \cite{Chung-RadiusMarginBound} full version
provided for \cite{Chung-RadiusMarginBound} writes ``For our current
implementation, solving each of $||w||^2$ ... $R^2$ ... is
considered an independent problem. In the future $||w||^2$ and $R^2$
... should be considered together. How to effectively pass
information under one given parameter set to another is also worthy
of investigation.''}; further below, we revisit this method and
elaborate.

2) A second bound, formed by the same product $R^2||\bw||^2$, is the
leave-one-out (loo) radius margin bound
\cite{Chung-RadiusMarginBound}
\begin{equation} \label{eqn:loo}
loo \leq 4R^2||\bw||^2
\end{equation}
which, \cite{Chung-RadiusMarginBound} stated holds for SVM without the bias term $b$ where $loo$ is the
number of loo errors, $\bw$ is the solution of (\ref{eqn:handle-sep}), and $R$ is the radius of the smallest
sphere containing all $\uphi(\bx)$. \cite{Chung-RadiusMarginBound} then stated that \cite{Vapnik-Chapelle}
``extends the bound for the general case where $b$ is present'' and also stated that ``it has been shown
(e.g. \cite{Vapnik_1998}) that $R^2$ is the objective value of the following optimization problem:''
\begin{equation} \label{eqn:radius}
\displaystyle\min_{\beta} 1 - \beta^{\rm T} K \beta \hspace{0.03in} s.t. \hspace{0.03in} 0 \leq \beta_n \forall n, \be^{\rm T} \beta = 1
\end{equation}
An alternative to (\ref{eqn:radius}) to estimate data radius is to
define $R^2$ as the maximum squared Euclidean distance between any
two (training) points: e.g. during 1-by-1 elimination of features,
$\max\limits_{i,j}||\bxi^{-m} - \bxj^{-m}||^2$. We elaborate on
these two different data radius formulations further below.

3) Furthermore, another publication endorsing utilization of data
radius for selecting among candidate functions was
\cite{Weston_NIPS} which, giving a theorem on the expectation of
error probability, wrote: ``This theorem justifies the idea that the
performance depends on the ratio $E\{R^2/M^2\}$ and not simply on
the large margin $M$, where $R$ is controlled by the mapping
function $\Phi(\cdot)$.'' Since the maximization of margin $M$ (or,
$1/||\bw||$, see e.g. \cite{Aksu_TNN}), central to SVM learning, is
commonly formulated via the minimization of WVSN as seen in the
hard-margin SVM formulation (\ref{eqn:handle-sep}), we can thus see
from \cite{Weston_NIPS} that performance can benefit from minimizing
(the expectation of) the product of squared radius and WVSN. As
discussed above in items 1 and 2, in the case of feature elimination
the computation of the two items in this product is straightforward
and computationally low-cost. Based on the importance of
radius-based bounding discussed somewhat briefly above, given also
that it was theoretically sound to propose for feature elimination
the LO (or MFE-LO) method \cite{Aksu_TNN} (that aims for, and
formulates, {\it margin maximization strictly in the hard-margin
sense} via minimizing the post-LO WVSN $(a^2)^{-\Mcal}
(||\bw||^2)^{-\Mcal}$; see: equation (12) in \cite{Aksu_TNN}), it is
also theoretically sound (once again {\it strictly in the
hard-margin sense}) to propose minimizing the product of $R^2$ and
the abovementioned WVSN quantity $a^2 ||\bw||^2$  i.e. ``the product
of squared radius and WVSN'' that we have been discussing above:
\begin{equation} \label{eqn:hbmfe-lo-general}
m^* = \displaystyle\mbox{arg}\min_{m \in \{\tilde{m}\in
\Rcal|g_l^{-\tilde{m}}>0 \forall l\}} \min_{a,b} (R^2)^{-m} (a^2)^{-m} (||\bw||^2)^{-m}
\end{equation}
Here, by moving $(R^2)^{-m}$ and $(||\bw||^2)^{-m}$
to the left by considering their values need not depend on the LO solver, (\ref{eqn:hbmfe-lo-general})
can be implemented as
\begin{equation} \label{eqn:hbmfe-lo}
m^* = \displaystyle\mbox{arg}\min_{m \in \{\tilde{m}\in
\Rcal|g_l^{-\tilde{m}}>0 \forall l\}} (R^2)^{-m} (||\bw||^2)^{-m} \min_{a,b} (a^2)^{-m}
\end{equation}
where the minimization on the right becomes easy to recognize as LO
(\ref{eqn:lo-lin}). Our novel method (\ref{eqn:hbmfe-lo-general})
(\ref{eqn:hbmfe-lo}) is named hBMFE-LO (``for hard-margin,
Bound-Based MFE-LO''), where the ``h'' emphasizes that this method
pertains to {\it margin in the hard-margin sense}; it is a method
that combines MFE (Margin-maximizing (or Margin-based) Feature
Elimination) \cite{Aksu_TNN} and upper bound on/for
misclassification risk. Fig. \ref{fig:sel1_mfelo-vs-mfeh}, discussed
earlier, illustrates hBMFE-LOe outperformed MFE-LO which does not
utilize radius. ``ER'' in the graph means the abovementioned
{\underline E}uclidean-based {\underline R} calculation, which is an
alternative to ``tR'' which means {\underline t}raining-based (or,
optimization-based) {\underline R} calculation such as given by the
optimization formulation (\ref{eqn:radius}). Extensive results for
several datasets and extensive discussion are given in Sec.
\ref{sec:results}.

\subsection{Utilizing radius in the soft-margin classifier sense} \label{sec:radius-soft}
Practicality of the soft-margin SVM was discussed as being
beneficial in past works and we cannot do justice to all of them
here; see e.g. some useful references for SVM mentioned in our
Introduction. \cite{Chung-RadiusMarginBound} mentioned that the
hard-margin SVM (\ref{eqn:handle-sep}) is ``not a form for practical
use. It may not be feasible if $\uphi(\bx)$ are not linearly
separable. In addition, a highly nonlinear $\uphi$ may lead to
overfitting.'' and mentioned next that thus practically they solve
the soft-margin SVM formulation e.g. (\ref{eqn:handle-nonsep}),
which they refer to as ``L1-SVM'' (where ``1'' in ``L1'' states the
exponent for the slackness variable in the objective function
(\ref{eqn:handle-nonsep}))\footnote{\cite{Chung-RadiusMarginBound}
also discusses and solves ``L2-SVM'' (for the case where the sum in
the soft-margin SVM objective function is instead
$\sum\limits_{n=1}^{N}\xi_n^2$) for which we do not perform
experiments herein.}. These comments, in support of practically
utilizing the soft-margin formulation (\ref{eqn:handle-nonsep})
instead of the hard-margin formulation (\ref{eqn:handle-sep}), are
complemented by the following abovementioned comments
\cite{Aksu_TNN}: ``Optimizing the amount of slackness (by choosing
the parameter $C$), {\it e.g.} via cross validation, may yield
classifiers with better generalization than those based on strictly
maximizing margin.''

Before continuing, we note an additional important information in
support of soft-margin SVM (i.e. in support of utilizing rather than
not utilizing slackness variables), that is, additional to the above
comments made in \cite{Chung-RadiusMarginBound} and elsewhere for
that support: slackness variables serve an important purpose even
when data is separable because, e.g., 1) as noted in
\cite{Aksu_TNN}, strictly satisfying the margin may lead to
overfitting when training samples at the margin are outliers or even
mislabeled samples, 2) classifiers that separate the data (i.e. with
zero classification error) and simultaneously allow some training
samples to lie within (i.e. violate) the margin can be obtained and,
due to reasons above, may generalize better than classifiers that do
not allow slackness (i.e. margin violation) when separating the data
with zero classification error. That is, in support of QP1, to
contrast QP1 to LO, we note that incorporating slackness variables
into the feature elimination model, as done by QP1, is important
{\it even when the data is expected to be separable}, as this can
alleviate overfitting, especially for data whose number of features
is very large because overfitting can be a cumulative effect that
stepwise accumulates over the course of elimination of a large
number of features such as during 1-by-1 elimination of features.

\begin{figure*}
\centering \subfigure[] {
   \includegraphics[scale=.7]{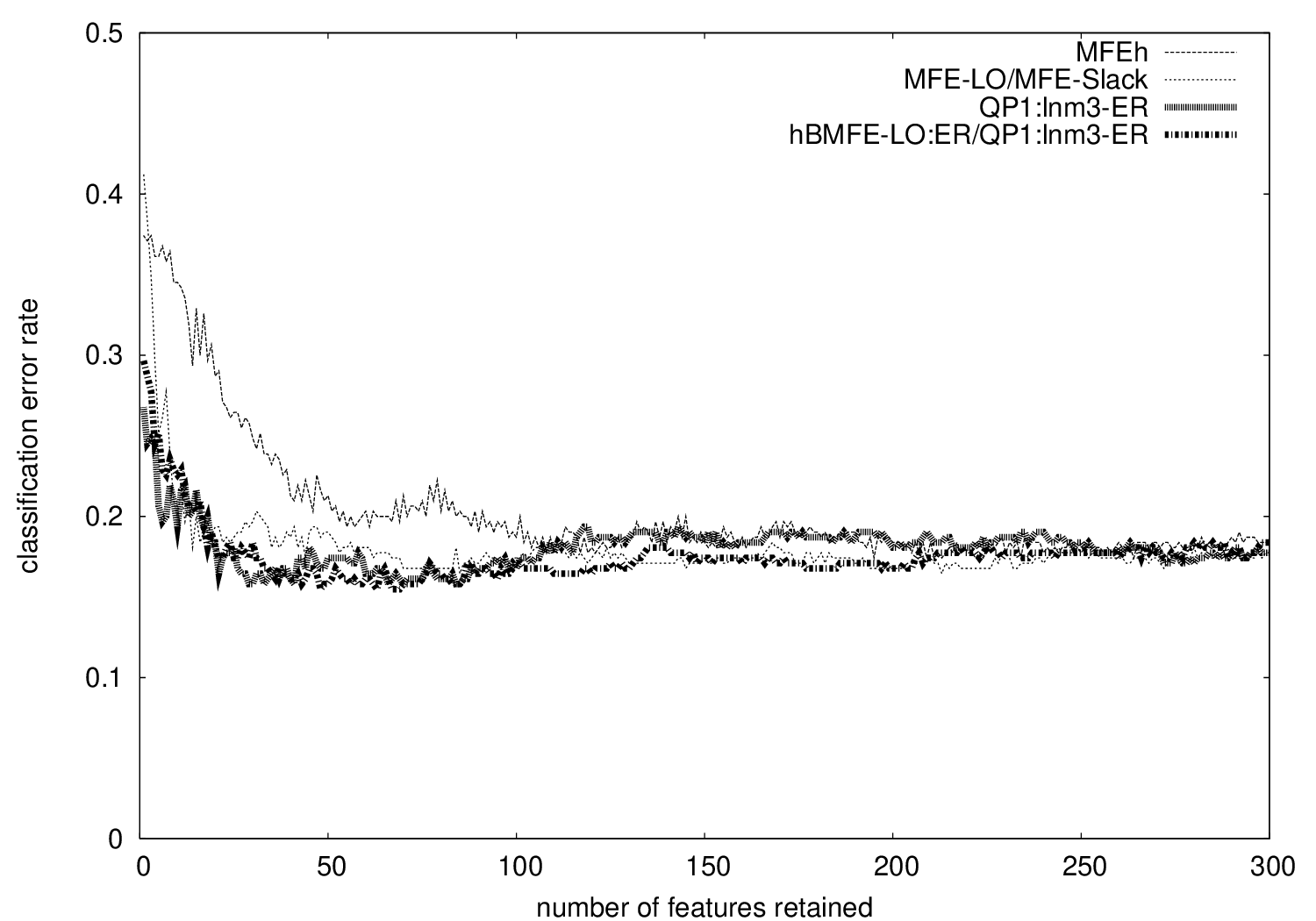}
} \caption[stuff]{QP1:lnm3-ER is placed into the earlier Figure
\ref{fig:sel1_mfelo-vs-mfeh} for comparison.}
\label{fig:sel1_qp1-vs-mfelo}
\end{figure*}

\begin{figure*}
\centering \subfigure[] {
   \includegraphics[scale=.6]{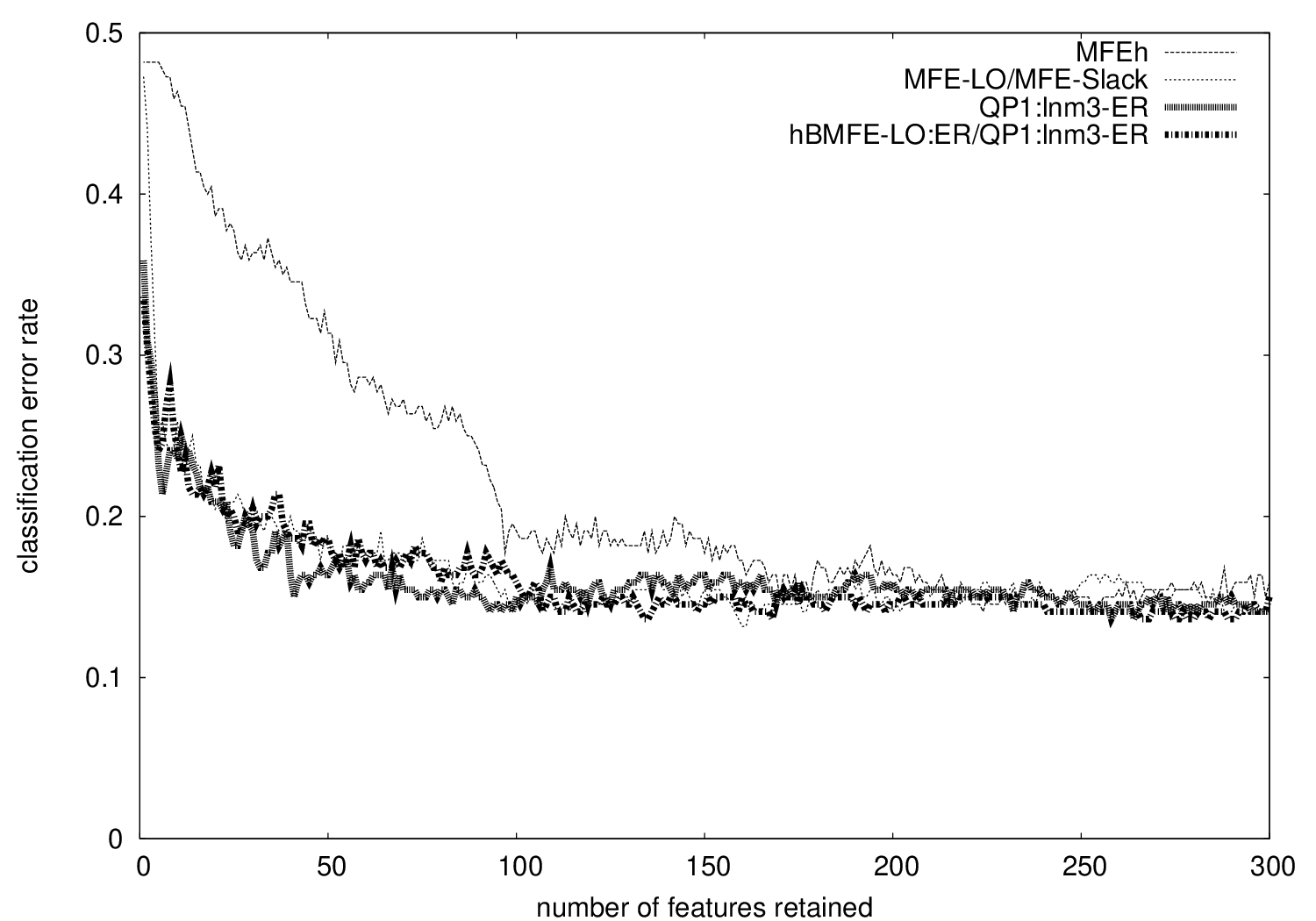}
} \caption[stuff]{Feature elimination starting with 7129 features.
Duke Breast Cancer gene dataset. SVM linear kernel case.}
\label{fig:sel1_qp1-vs-mfelo_duke}
\end{figure*}

\cite{Chung-RadiusMarginBound} stated that its goal is ``to make radius margin bound, a theoretical
bound of loo error, a practical tool'', and, based on its stated principle that ``finding a bound whose
minima are in a region with small loo values may be more important than its tightness'' it proposed
modified radius margin bounds for the soft-margin SVM (\ref{eqn:handle-nonsep}) (aka L1-SVM as mentioned earlier)
where, as \cite{Chung-RadiusMarginBound} states, ``the original bound is only applicable to the hard-margin case''. In particular, for the soft-margin SVM
case, \cite{Chung-RadiusMarginBound}
considered, and discussed its generated results for, the following three heuristic bounds for L1-SVM (\ref{eqn:handle-nonsep}):
\begin{equation} \label{eqn:3.1}
R^2 \be^{{\rm T}}\alpha + \sum\limits_{n=1}^{N}\xi_n
\end{equation}
\begin{equation} \label{eqn:3.2}
(R^2 + \frac{1}{C})(||\bw||^2 + C\sum\limits_{n=1}^{N}\xi_n)
\end{equation}
\begin{equation} \label{eqn:3.3}
(R^2 + \frac{\Delta}{C})(||\bw||^2 + 2C\sum\limits_{n=1}^{N}\xi_n)
\end{equation}
where $\Delta$ was considered to be a positive constant close to one
or one. For details in \cite{Chung-RadiusMarginBound}, we refer
interested readers to \cite{Chung-RadiusMarginBound}. In this
section, we focus on combining (\ref{eqn:3.3}) (using $\Delta=1$)
with our tunable QP1 optimization approach to define an accordingly
tunable novel feature elimination criterion that can potentially
achieve better generalization than 1) MFE-LO (which is non-tunable),
2) radius-incorporating hBMFE-LO proposed above (which is
non-tunable) and 3) eliminating using the QP1 criterion alone. That
is, the classifier we are interested to plug into the formulation
(\ref{eqn:3.3}) (i.e. values to plug in for the weight vector
squared norm and slackness values in (\ref{eqn:3.3})) is one that we
shall obtain via our QP1 approach. Since QP1 would perform best when
its tunability is utilized (by performing hyperparameter selection),
at each feature elimination step our new novel feature elimination
method, that combines QP1 with data radius utilization, performs
hyperparameter selection (since QP1 only takes little computation)
whereby many QP1 classifiers (i.e. $(a,b,\uxi)$ triplets) are
generated to select from, for that particular candidate feature
elimination; e.g. in the case of 1-by-1 elimination of features (and
when utilizing (\ref{eqn:3.3}) in particular), we thus propose the
following novel feature elimination method:
\begin{equation} \label{eqn:QP1-fe}
m^* = \displaystyle\mbox{arg}\min_{m\in \Rcal} \min_{j}((R^2)^{-m} +
\frac{1}{C_j})((w^2)^{-m} + 2C_j\sum\limits_{n=1}^{N}\xi_n^{-m})
\end{equation}

where $(w^2)^{-m}$ and $\xi_n^{-m}$ are the ``square of the scalar
weight $w$'' value and the scalar slackness $\xi_n$ values generated
by the QP1 training in the reduced space (i.e. when $m$ is the
candidate feature elimination being considered), and the set of
indices $j$ represents the set of hyperparameter value candidates.
Since there are multiple ways to generate a solution for QP1 as well
as generate the $R^2$ value, there are multiple ways to implement
the QP1-based feature elimination criterion given by
(\ref{eqn:QP1-fe}). The first way we discuss is named the
QP1:lnm3-ER method, where ``l'' means that {\underline L}IBSVM is
the means used by this particular way to generate a QP1 solution
(i.e. we train a 1d SVM using LIBSVM; we can, in future work,
alternatively train using our active-set method), ``n'' means we
make {\underline n}o modifications to LIBSVM's output for the 1d SVM
(C-SVC) training (i.e. no modifications to the set of positive
Lagrange multipliers assigned by LIBSVM and the samples they are
assigned to)\footnote{Our experience with LIBSVM 1d SVM training is
that sometimes the discriminant $yf$ function value is not equal to
$1$ for any of the abovementioned vectors being assigned positive
Lagrange multipliers.}, ``m'' means {\underline m}odel selection
(here aka hyperparameter selection) is performed, ``3'' means we
utilize the {\underline third} of the above three L1-SVM heuristic
bounds (i.e. bound (\ref{eqn:3.3})), and ``ER'' was discussed above.
In Fig. \ref{fig:sel1_qp1-vs-mfelo}, QP1:lnm3-ER is placed into the
earlier Figure to compare generalization performance with those
earlier methods. For a more potent illustration of the comparison of
QP1:lnm3-ER to other methods, we also give Fig.
\ref{fig:sel1_qp1-vs-mfelo_duke} wherein QP1:lnm3-ER essentially
outperforms hBMFE-LO:ER even though these two curves are, once
again, fluctuant; one would expect the QP1:lnm3-ER curve to become
even lower by simply expanding the search used for the stepwise
hyperparameter selection that QP1:lnm3-ER utilizes, by e.g. simply
including additional candidate $C$ values in the search set.

\section{Results and Discussion} \label{sec:results}
{\it Note from author Yaman M. Aksu: Some shortcomings of this
current particular version of this manuscript include the facts that
I did not yet have time to: 1) provide discussion of the results
that I am currently placing into this Results and Discussion section
(though they are somewhat self-explanatory), 2) provide isolated
(explicitly better organized) pseudocode for my novel specialized
active-set algorithm (in fact, algorithms, as there are variants)
that are discussed in the Appendix, 3) provide more discussion on
feature selection in especially the Introduction, 4) etc.}

The common procedure used for training an {\it initial} SVM
classifier, a first step for all feature elimination methods here,
randomly split the dataset 50-50\% into a non-heldout (training) set
$X$ and a heldout (test) set $\bar{X}$ (with each split defining one
`trial'), selected hyperparameters by 5-fold cross-validation
\cite{Duda} on $X$, and used all of $X$ to retrain the trial's
classifier for these selected values. In Figures we show
across-trial averages. When features outnumber samples (e.g. $7129
\gg$ tens or hundreds), e.g. in gene, biomedical image, and other
domains, it is highly probable that the training set will be
separable while eliminating all the way down to {\it relatively few}
features (e.g. hundreds, tens) \cite{Cover_ITEC,Aksu_TNN}, and thus
all methods herein may be able to eliminate all the way down to {\it
relatively few} features without losing separability, whereas for
intermediate dimensioned data separability may be lost {\it sooner}
e.g. when {\it half} of the features is still left to eliminate.

For QP1, as mentioned above, at each feature elimination step we
performed hyperparameter selection, to select from a set of
candidate $C_j$ values (\ref{eqn:QP1-fe}); the set was
$\{C_{init},C_{init}/2^{1},\ldots,C_{init}/2^{30}\}$ where
$C_{init}$ denotes the $C$ value that was used for training the
initial SVM classifier (chosen in full feature space, by 5-fold
cross-validation) prior to the feature elimination
process\footnote{Except the Leukemia dataset, for which a more
balanced choice of $C$ values was made wherein not only values
smaller than $C_{init}$ but also larger than $C_{init}$ were
included:
$\{C_{init}\cdot2^{5},\ldots,C_{init}\cdot2^{1},C_{init},C_{init}/2^{1},\ldots,C_{init}/2^{15}\}$.}.
Alternatively, in future work, training the initial classifier and
performing feature elimination can be carried out jointly rather
than separately, to jointly incorporate hyperparameter selection.
\begin{figure*}
\centering \subfigure[Zoomed to 300 features retained starting with
7129 features.] { \label{fig:duke_300}
    \includegraphics[scale=.6]{apr22_6_graph_zoomed300selector1_duke_kernel0.eps}
} \subfigure[Zoomed to 600 features retained starting with 7129
features.] {
    \includegraphics[scale=.6]{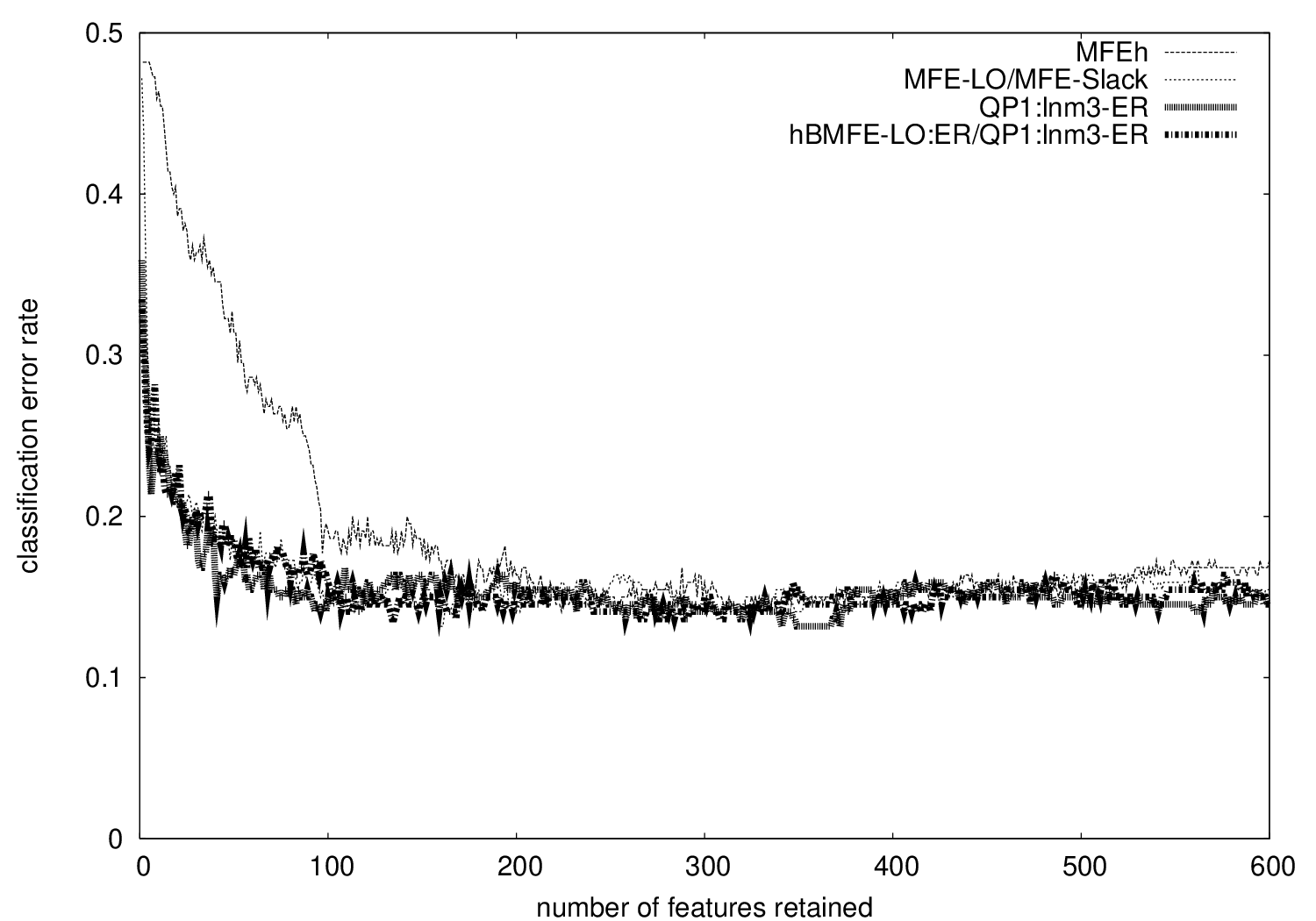}
}\caption[]{Average test set classification error rate for the Duke
Breast Cancer gene dataset with 7129 features and much fewer
samples. SVM linear kernel case.} \label{fig:duke}
\end{figure*}
\begin{figure*}
\centering \subfigure[] {
    \includegraphics[scale=.6]{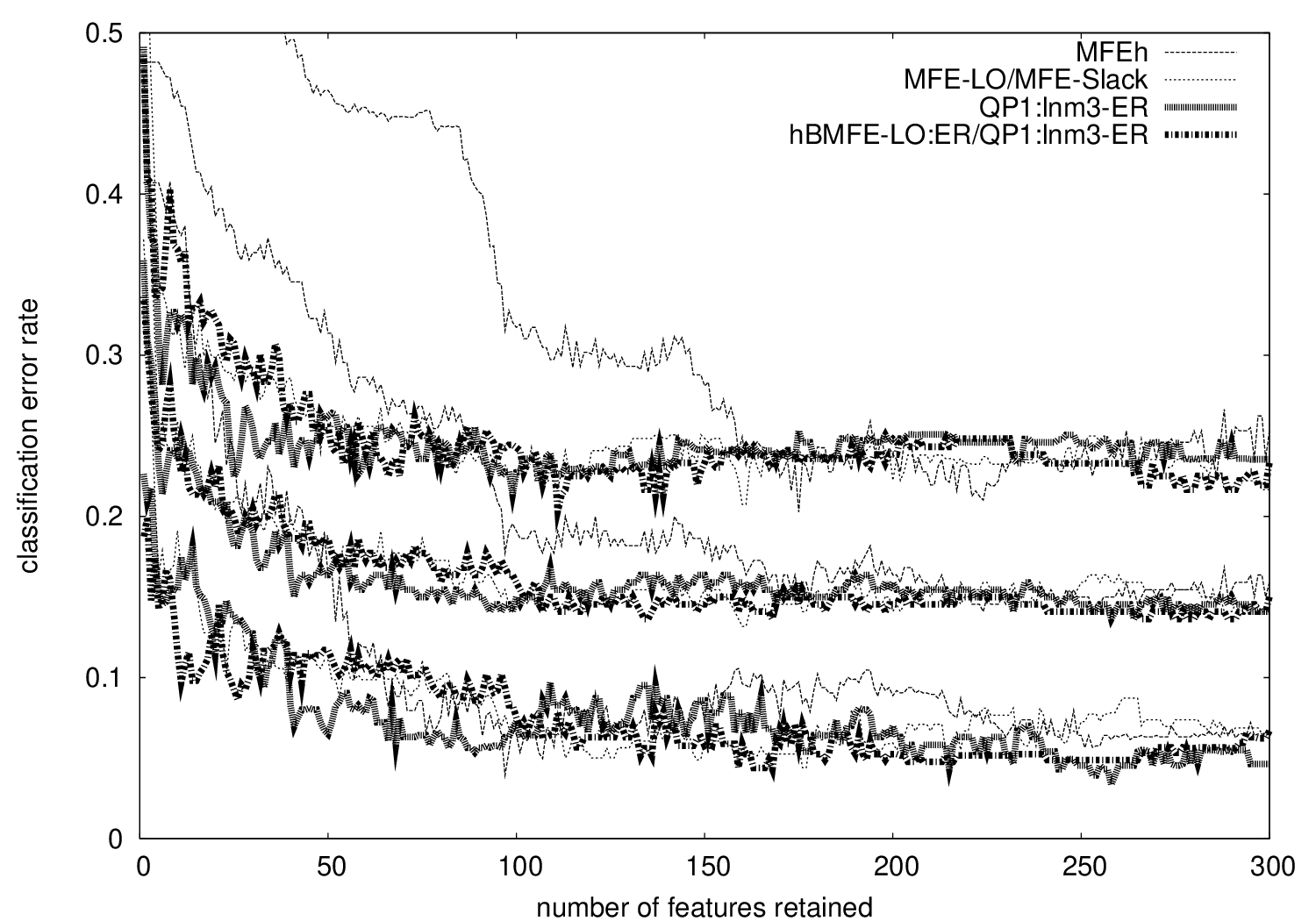}
} \caption[]{This Figure redraws Fig. \ref{fig:duke_300} (i.e. the
across-trial average $\mu$) so as to also include, for each
elimination method, the $\mu+\sigma$ curve (seen above the average
curve $\mu$) and the $\mu-\sigma$ curve (seen below the average
curve $\mu$), where $\sigma$ is the across-trial standard deviation
of the elimination method.} \label{fig:duke_300_enhanced}
\end{figure*}
\begin{figure*}
\centering \subfigure[Zoomed to 300 features retained starting with
7129 features.] { \label{fig:leu_300}
    \includegraphics[scale=.7]{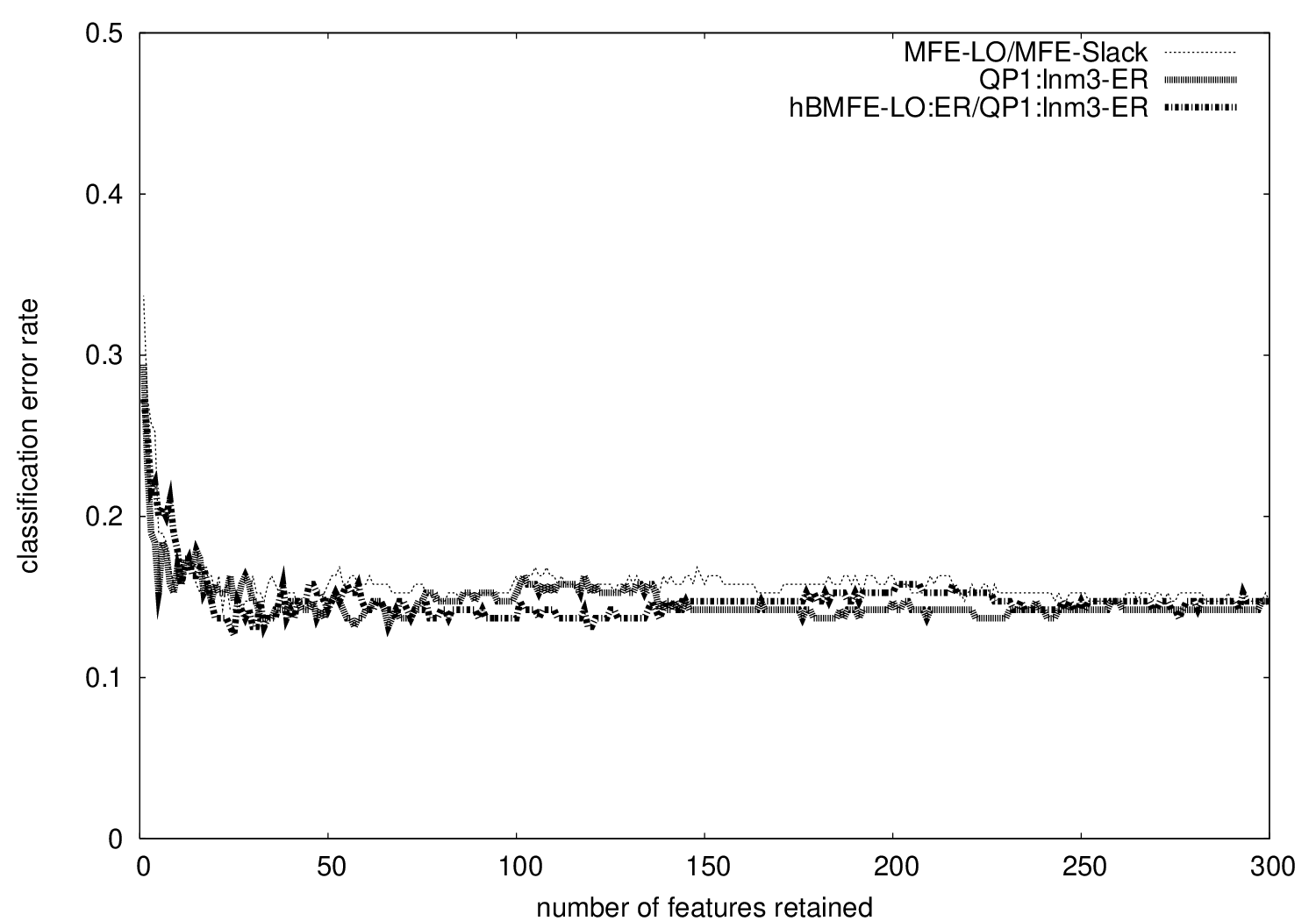}
} \subfigure[Zoomed to 600 features retained starting with 7129
features.] {
    \includegraphics[scale=.6]{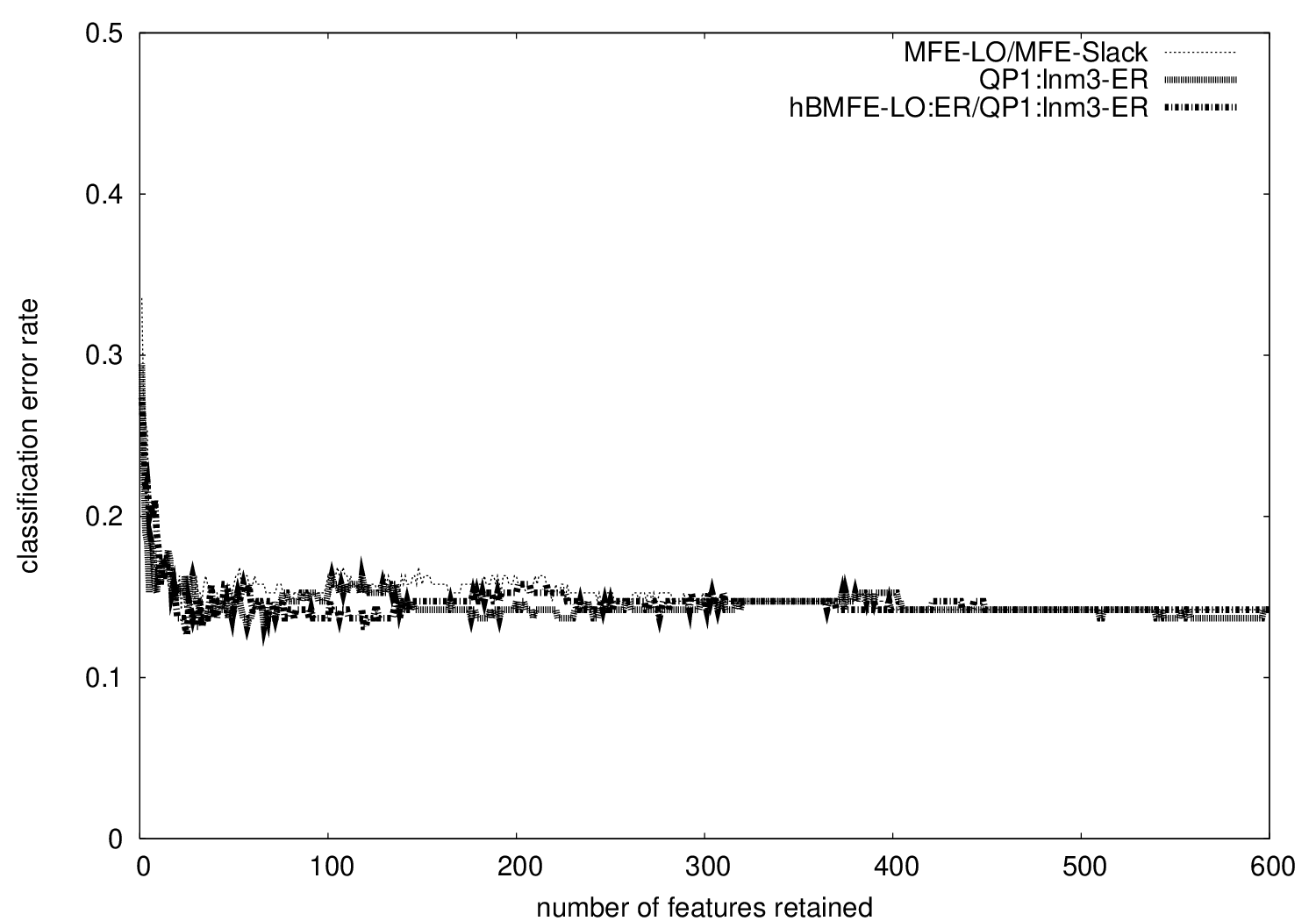}
}\caption[]{Average test set classification error rate for the
Leukemia gene dataset with 7129 features and much fewer samples. SVM
linear kernel case.}
\end{figure*}
\begin{figure*}
\centering \subfigure[] {
    \includegraphics[scale=.7]{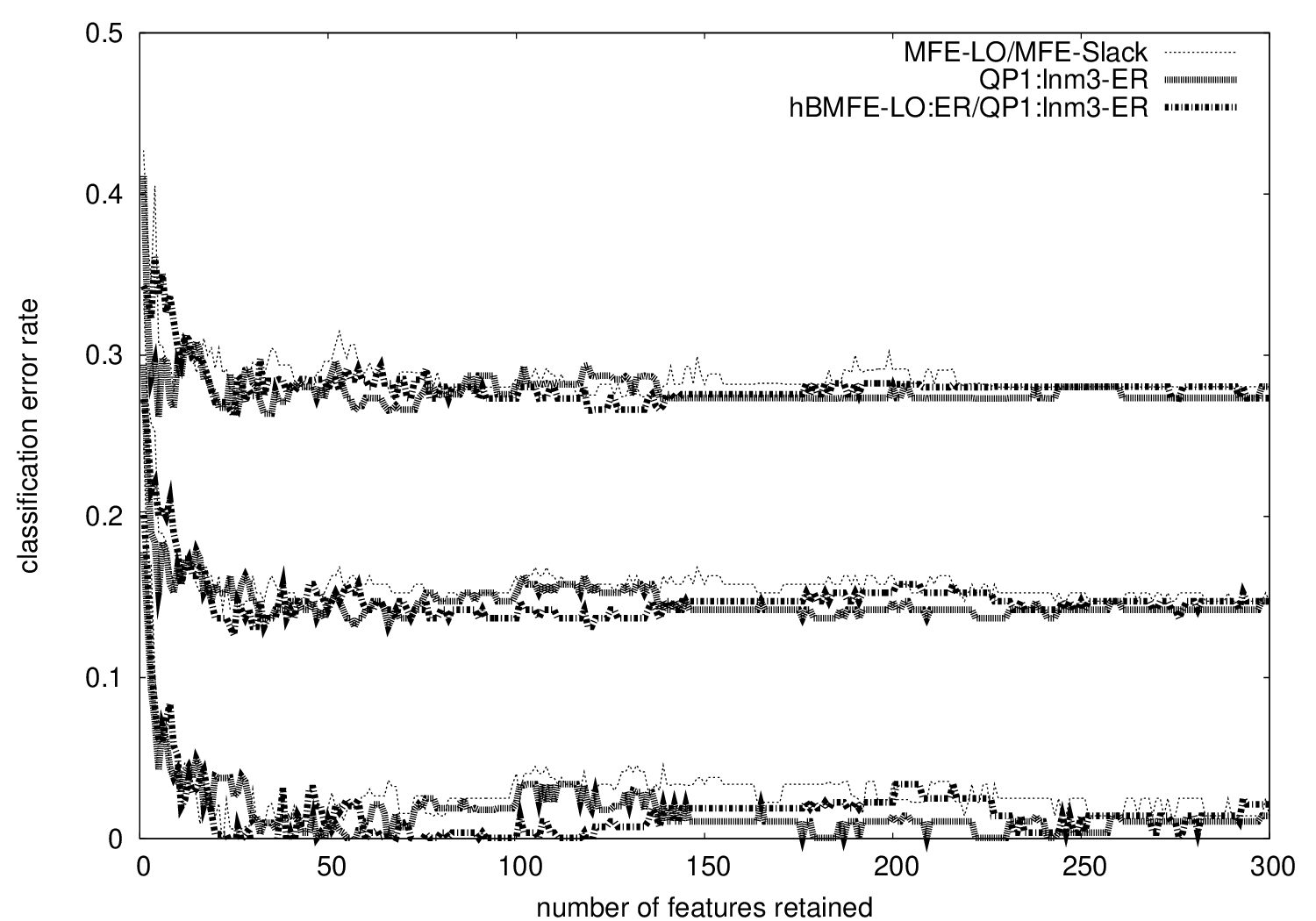}
} \caption[]{This Figure redraws Fig. \ref{fig:leu_300} (i.e. the
across-trial average $\mu$) so as to also include, for each
elimination method, the $\mu+\sigma$ curve (seen above the average
curve $\mu$) and the $\mu-\sigma$ curve (seen below the average
curve $\mu$), where $\sigma$ is the across-trial standard deviation
of the elimination method.}
\end{figure*}
\begin{figure*}
\centering \subfigure[Average test set classification error rate for
the Colon Cancer gene dataset with 2000 features and much fewer
samples. SVM linear kernel case.] { \label{fig:test-colon}
    \includegraphics[scale=.6]{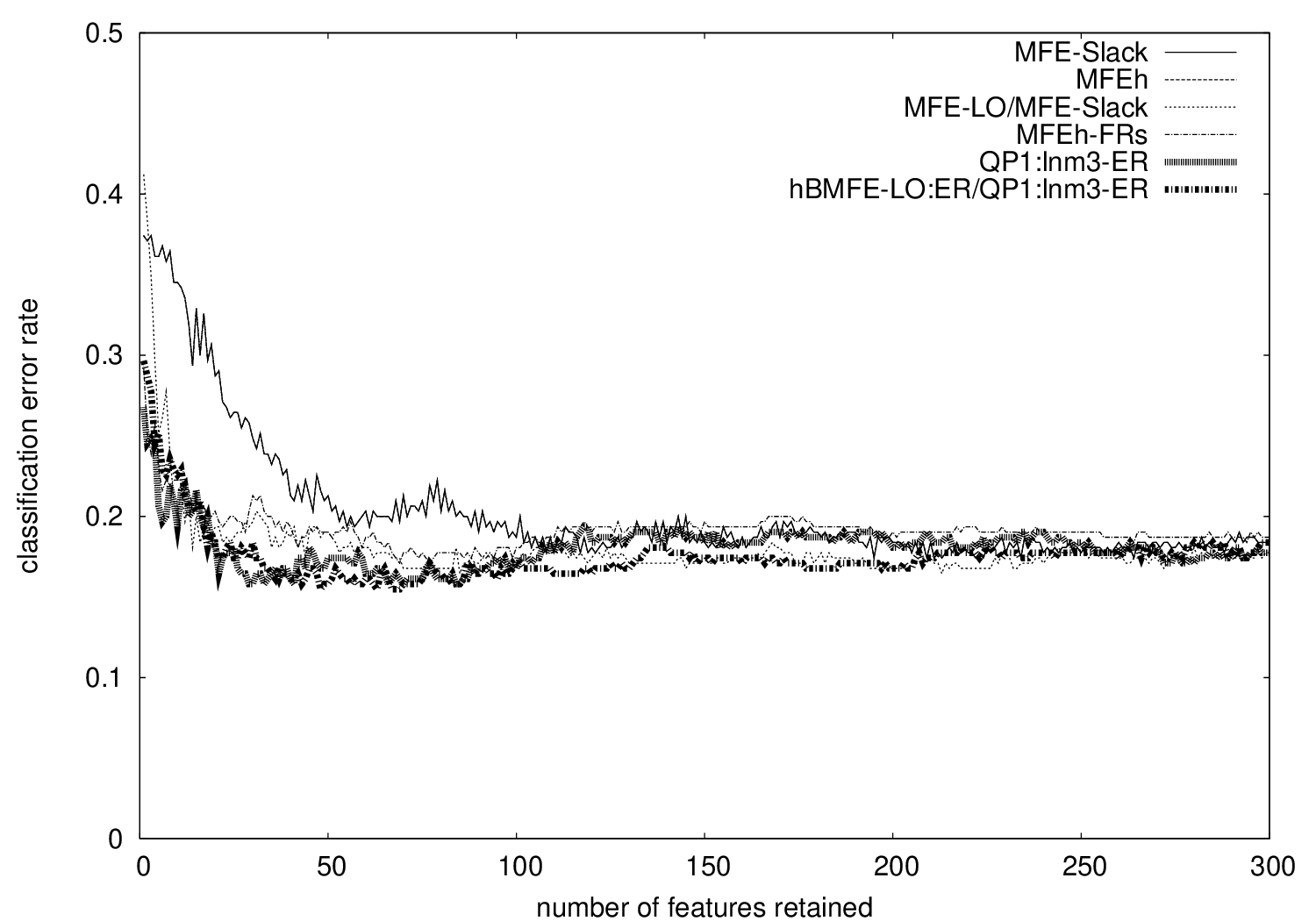}
} \subfigure[This Figure redraws Fig. \ref{fig:test-colon} (i.e. the
across-trial average $\mu$) so as to also include, for each
elimination method, the $\mu+\sigma$ curve (seen above the average
curve $\mu$) and the $\mu-\sigma$ curve (seen below the average
curve $\mu$), where $\sigma$ is the across-trial standard deviation
of the elimination method.] {
    \includegraphics[scale=.7]{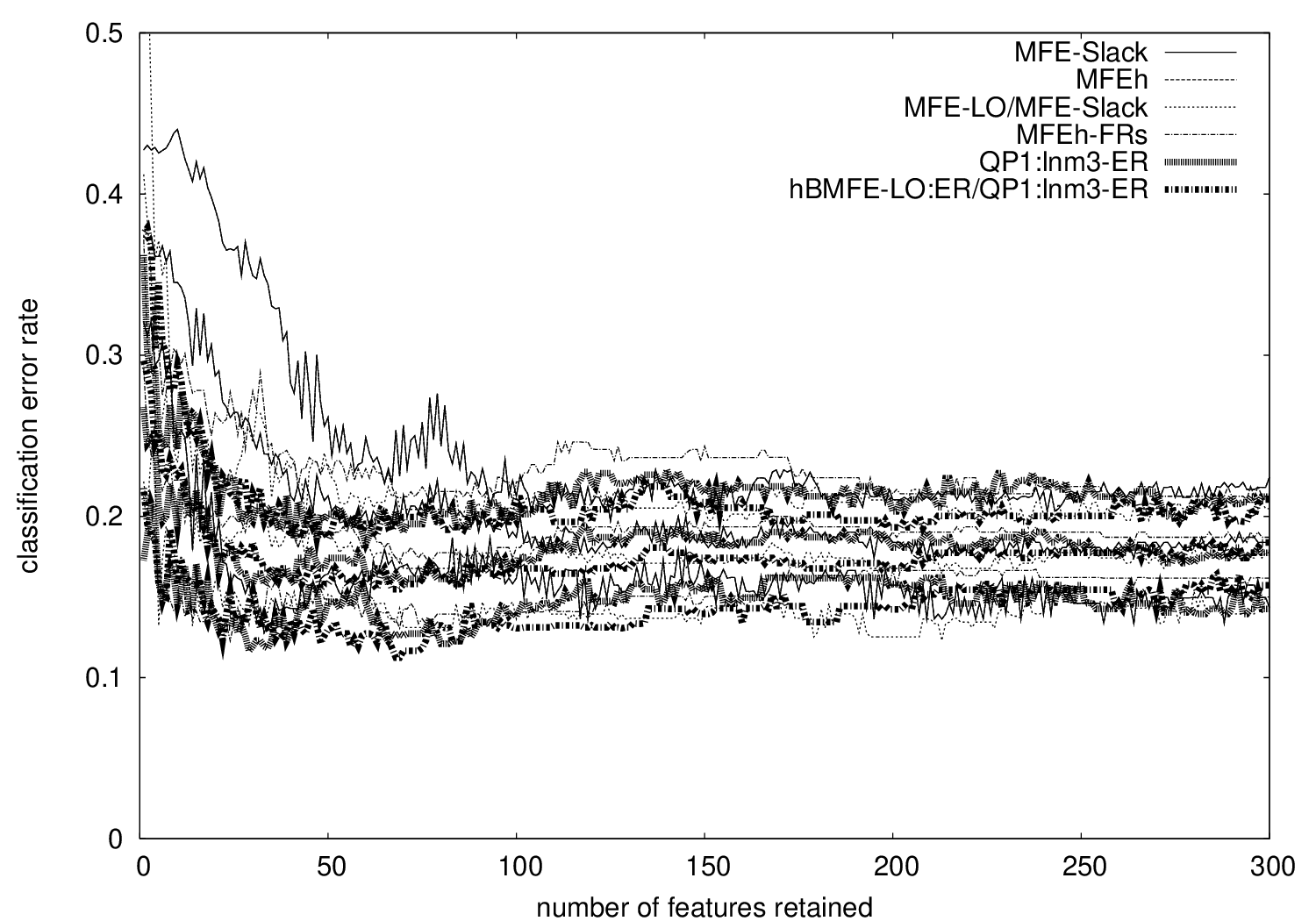}
} \caption[]{} \label{fig:colon}
\end{figure*}
\begin{figure*}
\centering \subfigure[Average test set classification error rate for
the Splice Scale dataset with 60 features; a dataset for the
low-to-intermediate number of features case. SVM linear kernel
case.] {
    \label{fig:test-splice}
    \includegraphics[scale=.5]{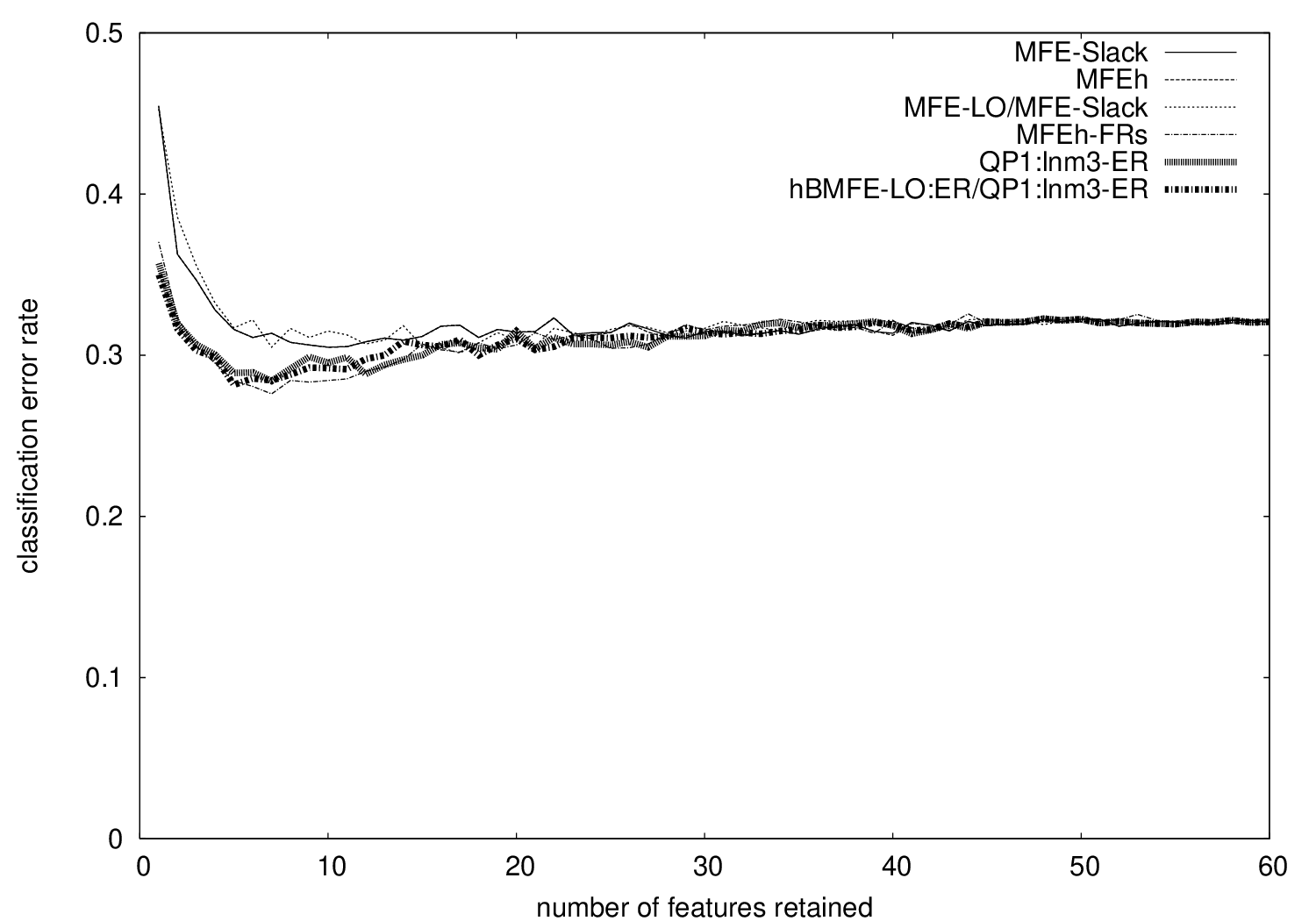}
} \subfigure[This Figure redraws Fig. \ref{fig:test-splice} (i.e.
the across-trial average $\mu$) so as to also include, for each
elimination method, the $\mu+\sigma$ curve (seen above the average
curve $\mu$) and the $\mu-\sigma$ curve (seen below the average
curve $\mu$), where $\sigma$ is the across-trial standard deviation
of the elimination method.] {
    \includegraphics[scale=.5]{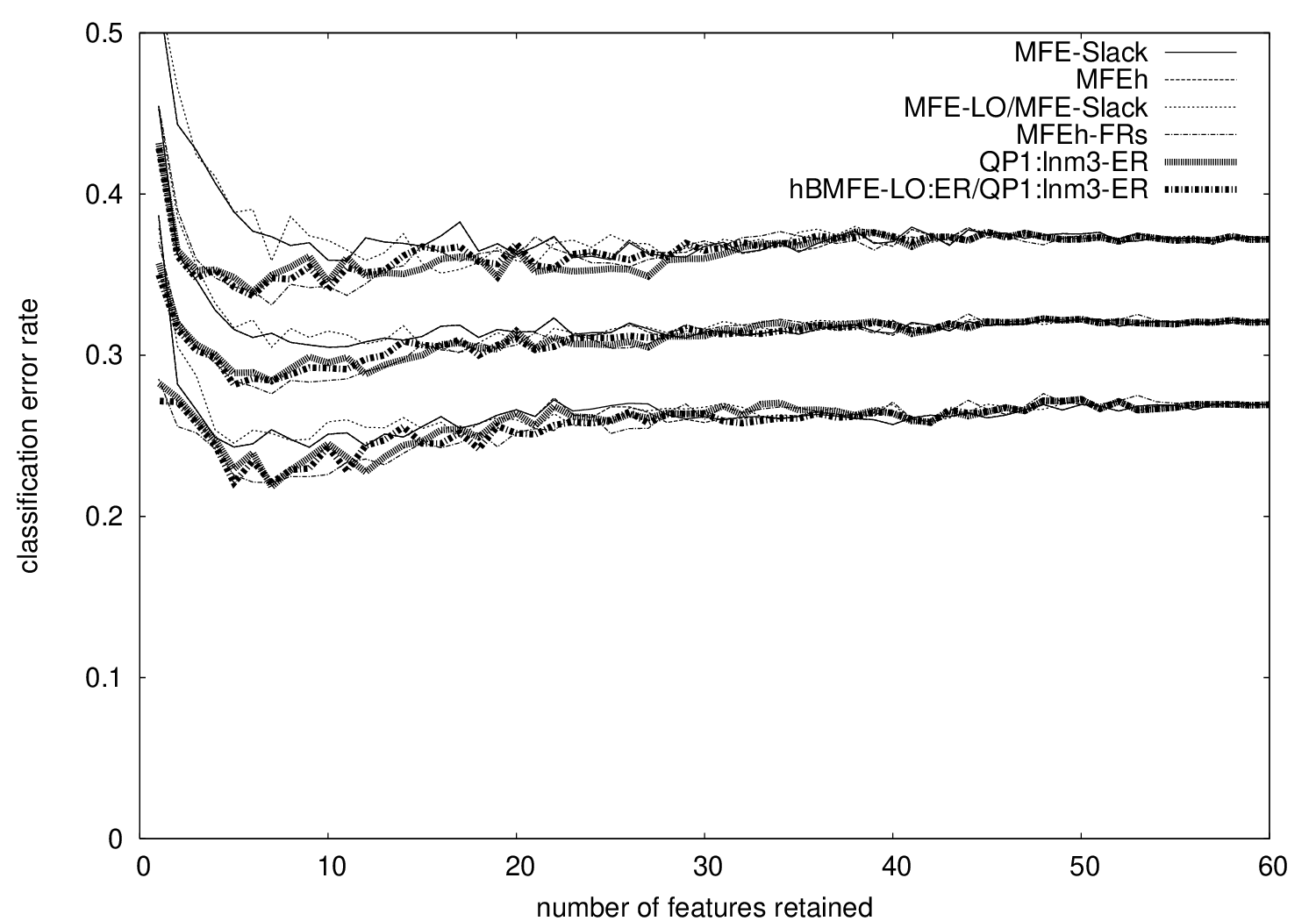}
} \caption[]{} \label{fig:splice}
\end{figure*}

\section{Appendix 1: QP1-specialized computationally low-cost active-set method} \label{sec:descent}

\subsection{Introduction} \label{sec:descent:intro}

Approaching QP1 as an inequality constrained quadratic programming
(ICQP) problem in primal form, we specialized an active-set method
whereby we make this method of obtaining a QP1 solution
computationally low-cost as well. This method of obtaining a QP1
solution, which considers QP1 in primal form (i.e. takes descent
steps to directly minimize (the primal form of) the QP1 objective
function), is an alternative to considering QP1 a 1d SVM problem to
be solved by an SVM solver in dual form.

In this introductory subsection, for the reader we provide an
informative summary, based on \cite{Nocedal}, about how an ICQP
problem
\begin{equation}
\label{eqn:icqp} \bqstar = \min_{\bq} \frac{1}{2} \bq^{{\rm T}} \bG
\bq + \bd^{{\rm T}}\bq \hspace{0.1in} s.t. \hspace{0.05in}
\bai^{{\rm T}}\bq \geq t_i, \hspace{0.05in} i\in I
\end{equation}
can be solved via an active-set method; additional detail can be
found in e.g. \cite{Nocedal}. The notation in (\ref{eqn:icqp}) is a
standard one.\footnote{Shortly Sec. \ref{sec:descent:our} will state
these variables (e.g. $\bq$, $\bG$, $\bd$, $\bai$) for our
particular QP problem QP1.} Then, after the Introduction, in Sec.
\ref{sec:descent:our}, we focus on the forms, and properties, of
these particular QP1-specific variables and matrices (e.g. $\bq$,
$\bG$, $\bd$, $\bai$)\footnote{Notice the matrices at hand here in
QP1 exhibit much regularity i.e. contain numerous zeros and ones and
we will make use of this in Sec. \ref{sec:descent:our}; e.g. $\bG$
has a single nonzero element, and almost all of $\bA$ and all of $\bd$
are zeros and ones.}, devise (given those particular matrices) our
e.g. Lemmas and Theorem i.e. our mathematical contributions for
finding a QP1 solution (especially, a computationally low-cost
solution), and accordingly devise an active-set algorithm
specialized for QP1 in particular that obtains a computationally
low-cost solution for QP1.

For small- to medium-scale ICQP problems, it has been mentioned that
active-set methods are the most effective \cite{Nocedal} generally;
we took a specific (and yet fairly general, as discussed below)
active-set algorithm provided by \cite{Nocedal} and specialized it
in two central ways. The first specialization is that our work
addresses an important matter about positive definiteness
(pertaining to solving QP problems via an active-set
approach\footnote{Specifically, the positive definiteness of
$\bZ^{{\rm T}} \bG \bZ$, a matrix we discuss below.}) in two
different ways\footnote{The first way will require $\bZ^{{\rm T}}
\bG \bZ$ to be positive definite; the second way will not.}, even though
it seems \cite{Nocedal} did not discuss the algorithm separately for
these two separate ways in conjunction with presenting the
algorithm.\footnote{Our work is, of course, helped by the fact that
the QP problem has a more specific form than the general form
(\ref{eqn:icqp}) used by \cite{Nocedal}, as in our case the matrices that appear in the
problem definition have a specific, known form, as we shortly
discuss.} Our second specialization of the algorithm in
\cite{Nocedal} is of course that we focus on our particular novel
quadratic programming (QP) problem statement (\ref{eqn:qp-linear})
(i.e. our QP1 is a specific QP problem, defined by specific matrix
forms) and create a computationally low-cost algorithm in
doing so. Herein we present our active-set
method work as a novel theoretical mathematical contribution
(wherein our devised Lemmas and Theorems are presented) even though
we do not actually utilize this algorithm in our current feature elimination
experiments herein.

Usually a primal iterative active-set method starts with a feasible
initial $\bqzero$ and ensures each $\bqk$ (at iteration $k$) is
feasible \cite{Nocedal}. An optimal active set (the active set for
$\bqstar$)\footnote{The active set $\Acal(\bq)$, at some feasible
$\bq$, identifies constraints $i$ fulfilled as equalities at $\bq$
i.e. $\bai^{{\rm T}}\bq = 0$ \cite{Nocedal}.} is sought, via such
iterations, at each of which, one constraint is dropped or added to
the current (iteration's) estimate of this set, called the Working
Set $W$ ($W^k$ at iteration $k$)\footnote{Notice that by definition
of W a constraint may be active without being in W.} \cite{Nocedal}.
Specifically, to ensure $\bqkplusone$ is feasible, the direction
$\bpkstar$, along which to move from $\bqk$ to reach $\bqkplusone$
(i.e. $\bqkplusone = \bqk + \delta_k \bpkstar$ for some $\delta_k
\in \mathbb{R}$) is computed such that constraints identified by a
set $W^k$ are fulfilled as equalities i.e. $\bai^{{\rm T}}\bpkstar=0
\hspace{0.05in} \forall i \in W^k$; this ensures feasibility of $\bqkplusone$
because these $W^k$ constraints are also fulfilled at $\bqkplusone$ due to
$\bai^{{\rm T}}\bqkplusone = \bai^{{\rm T}}(\bqk + \delta
\bpkstar)=\bai^{{\rm T}}\bqk=t_i$, for any $\delta$ \cite{Nocedal}.
I.e., this is an equality-constrained QP (ECQP) subproblem (of the
ICQP problem), with the above constraint set $W^k$:
\begin{equation}
\label{eqn:ecqp} \bpkstar = \min_{\bpk} \frac{1}{2} \bpk^{{\rm T}}
\bG \bpk + \bhk^{{\rm T}}\bpk \hspace{0.1in} s.t. \hspace{0.1in}
\bai^{{\rm T}}\bpk=0 \hspace{0.05in} \forall i \in W^k
\end{equation}
where $\bhk \equiv \bG \bqk + \bd$ which can be evaluated prior to
solving (\ref{eqn:ecqp}) \cite{Nocedal}. Let $\bA$ denote the matrix
with {\it i}-th row $\bai^{{\rm T}}$. Let $\bAI$ denote the matrix
whose only rows are a subset of $\bA$'s as specified by an index set
$\Ical$; e.g. $\bAWk$ denotes the matrix whose only rows are
$\bai^{{\rm T}}$ for constraints $i \in W^k$ in (\ref{eqn:ecqp}).

Let $N(\bAW)$ denote the null space of $\bAW$; a complete set of
basis vectors for $N(\bAW)$ can be arranged as columns of a matrix,
denoted $\bZ$ herein. By Lemma 16.1 in \cite{Nocedal}, the
first-order necessary conditions for $\bpkstar$ to be a solution of
(\ref{eqn:ecqp}) can be fulfilled in conjunction with requiring the
KKT matrix $\left(
  \begin{array}{cc}
    \bG & -\bAW^{{\rm T}} \\
    \bAW & \bzero \\
  \end{array}
\right)$ to be nonsingular \cite{Nocedal}, by requiring $\bAW$ to
have full row rank and assuming that $\bZ^{{\rm T}} \bG \bZ$ is
positive definite; that is, by making the following two assumptions
(requirements):

{\it R1:} Require rows of $\bAW$ to be linearly
independent.\footnote{That is, LICQ (Linear Independence Constraint
Qualification) is fulfilled for active constraint gradients
\cite{Nocedal}, whereby valid use of KKT conditions (to solve the
constrained optimization problem at hand) is enabled.}

{\it R2:} Require $\bZ^{{\rm T}} \bG \bZ$ to be positive definite.

\cite{Nocedal} gives an active-set algorithm, ``Algorithm 16.1
(Active-Set Method for Convex QP)'' (aka A16.1 herein). Since herein
we utilize this algorithm, we now discuss it in the context of {\it
R1} and {\it R2}:

{\it For R2:} For solving the ECQP, the A16.1 algorithm only says
``solve'' and also a discussion of whether {\it R2} shall be
fulfilled does not seem to be provided in \cite{Nocedal} in
conjunction with presenting the algorithm. Nevertheless, when
applying A16.1 to our particular QP problem (\ref{eqn:qp-linear}) in
Sec. \ref{sec:descent:our} we provide a way that focuses on the
prospect of ensuring that {\it R2} is fulfilled, due to the above
Lemma 16.1 remark about the solution $\bpkstar$ of the
ECQP.\footnote{Note that {\it R2} is for the case $N(\bAWk)\neq
\emptyset$ (i.e. the null space does not only contain the
zero-vector, i.e. $\bZk\neq \bzero$); below we will additionally
address the possibility of $N(\bAWk) = \emptyset$ (i.e. the null
space contains only the zero-vector i.e. is ``empty'', i.e.
$\bZk=\bzero$).}

{\it For R1:} (This paragraph too gives mathematical details that
pertain to the perspective that the problem is an ICQP problem,
rather than pertain to the perspective that the problem also happens
to be an SVM problem; the reason we do not here yet intuitively
associate these mathematical details with SVM is that such
association is postponed to Sec. \ref{sec:descent:our}, as mentioned
earlier.) The strategy of A16.1 for {\it R1} to be fulfilled at
every ICQP iteration $k$ is to start (at $k=0$) with an $\bAWzero$
that fulfills {\it R1} and to shrink or grow $\bAW$ by at most a
single row of $\bA$ (i.e. a single constraint of the ICQP) at each
iteration (if not keeping $\bAW$ the same) while ensuring the row
chosen to grow $\bAW$ is linearly independent of the existing rows
(of $\bAW$). Specifically, in the event $\bpkstar$ is found to be
nonzero, denoting $B\equiv\{i\notin W^k| \bai^{{\rm
T}}\bpkstar<0\}$, in A16.1 \cite{Nocedal} the ratio
$R_i\equiv(t_i-\bai^{{\rm T}}\bqk)/\bai^{{\rm T}}\bpkstar$ is
computed for each $i\in B$, so as to compute $\delta_k\equiv
\min_{i\in B}(1,R_i)$ and $j\equiv \mbox{arg}\min_{i\in B}(1,R_i)$,
so that, accordingly, if $\delta_k<1$, $W^{k+1}$ is set to $W^k \cup
\{j\}$, with $j$ referred to as the ``blocking constraint''. It is a
``blocking constraint'' because, as can be easily seen from the
definitions of $B$ and $R_i$ that pertain to the abovementioned
movement along the direction $\bpkstar$, taking along that
$\bpkstar$ direction a whole step $1\bpkstar$ (i.e. $\delta_k
\bpkstar$ for $\delta_k=1$) is being blocked by the fact that one of
the constraints (i.e. a ``blocking constraint'') is becoming active
upon traveling merely a fraction $\delta_k<1$ of that whole step;
the directional distance traveled is thus $\delta_k \bpkstar$ where
$\delta_k<1$. Else if a whole step can be travelled (i.e. $\delta_k$
is $1$), $W^{k+1}$ is set to $W^k$; i.e., without having to modify
the Working Set, we have moved an amount $\delta_k \bpkstar$ from
$\bqk$ and arrived the new location $\bqkplusone$. Else in the event
$\bpkstar$ is instead found to be zero, A16.1 states it has reached
its terminating condition unless the Lagrange multiplier for a
constraint $i\in W^k$ was found to be negative in which case A16.1
sets $W^{k+1}$ to $W^k\setminus i$ \cite{Nocedal}; i.e., A16.1
removes constraint $i$ from the Working Set since by convention an
initial global assumption requiring Lagrange multipliers to be
nonnegative was made (as often is made when utilizing Lagrange
multipliers).

\subsection{Specializing Algorithm A16.1 \cite{Nocedal} to our particular QP problem QP1} \label{sec:descent:our}
By comparing (\ref{eqn:qp-linear}) to (\ref{eqn:icqp}), notice in
our QP1 formulation (\ref{eqn:qp-linear}) that $I =\{1,\ldots,2N\}$,
$\bq \equiv [\bqonetwo^{{\rm T}} \hspace{0.05in} \xi_{1}
\ldotswithhspace \xi_{N}]^{{\rm T}}$ where $\bqonetwo \equiv [a
\hspace{0.05in} b]^{{\rm T}}$, $\bG \equiv \left(
  \begin{array}{cc}
    (||\bw||^2)^{-\Mcal} & \bzero_{1 \times N+1} \\
    \bzero_{N+1 \times 1} & \bzero_{N+1 \times N+1} \\
  \end{array}
\right)$, $\bd\equiv [0 \hspace{0.05in} 0 \hspace{0.05in}
C\bone_{1\times N}]^{{\rm T}}$, $\bai^{{\rm T}}$ is the {\it i}-th
row of $\bA \equiv \left(
  \begin{array}{cc}
    \bV_{N\times 2} & \bI_{N\times N} \\
    \bzero_{N\times 2} & \bI_{N\times N} \\
  \end{array}
\right)$, $\bV_{N\times 2}$ consists of $1\times 2$ rows
$\bvone,\ldotswithhspace,\bvN$ where $\bvn\equiv[y_n
(\bw^{-\Mcal})^{{\rm T}} \bxn^{-\Mcal} \hspace{0.1in} y_n]$,
$\by\equiv [y_1 \ldotswithhspace y_N]^{{\rm T}}$, $\bt \equiv [t_1
\ldotswithhspace t_{2N}]^{{\rm T}} \equiv [\bone_{1\times N}
\hspace{0.05in} \bzero_{1\times N}]^{{\rm T}}$. As mentioned earlier,
these matrices exhibit a specific, highly regular form defined by
e.g. many zeros and ones in fixed spots.

In discussing the properties of ECQP (note: not ICQP),
\cite{Nocedal} made the assumption (see: page 444) that in the ECQP
the number of constraints is not greater than the number of unknowns
(i.e. the number of optimization parameters). We shall refer to this restriction
as Restriction 1. When the ECQP occurs
within an active-set algorithm such as we discussed when giving
(\ref{eqn:ecqp}), this means the assumption that the number of
elements in the Working Set $W$ is not greater than the number of
optimization parameters. Accordingly, we now make the observation
that the active-set algorithm A16.1 in \cite{Nocedal} would be
suitable for QP1 if the number of elements in set $W$ (or,
equivalently, the number of rows in $\bAW$) is ensured to not be
greater than $N+2$ which is the number of parameters in the
parameter vector $(a,b,\xi_1,\ldots,\xi_N)$ of QP1. To
mathematically appreciate the above assumption made
by \cite{Nocedal}, one can consider it from the LICQ perspective, in
conjunction with $R1$ above, as follows. As mentioned above, to
utilize LICQ when solving the optimization problem at hand, one can
fulfill $R1$, but since $R1$ cannot be fulfilled in the
event $\bAW$ has more rows than columns (i.e. a simple fact from
linear algebra), $\bAW$
needs to be have fewer rows than columns to fulfill $R1$ and LICQ,
and this leads us back to the abovementioned assumption in
\cite{Nocedal}.

Shortly we will return to discussing Restriction 1. Now, let us
introduce a central point, a point that will be soon concluded by
our Lemmas and Theorem; this introduction, before those mathematical
details enter the picture, is to highlight this central point with
an intuitive and less mathematical description. The central point is
that in our specialization of A16.1 currently our focus when
calculating the step direction $\bpkstar$, and taking the step
$\delta_k \bpkstar$ from $\bqk$ to $\bqkplusone$, is {\it i)} to
ensure a sample is a ``doubly-active'' sample at $\bqk$ i.e. a
sample whose {\it both constraints} are active at $\bqk$ and {\it
ii)} find the direction $\bpkstar$ that both decreases the objective
function and keeps that sample doubly-active upon taking the step
$\delta_k \bpkstar$, and thus we refer to $\bpkstar$ as the sample's
``direction of remaining doubly-active (DRD)''. While moving along
that sample's DRD, a second sample can become doubly-active before a
whole step $1\bpkstar$ is completed, blocking further movement along
that DRD (whereby, the computed $\delta_k$ is less than $1$ and
reflects the amount of uninterrupted unblocked movement), in which
case the next movement can take place along that {\it second
doubly-active sample's DRD} (which would likewise be found by
transferring the ``doubly-active sample'' designation to {\it
solely} this new sample, just like that designation was previously
given to a single sample (the previous sample) in calculating the
direction $\bpkstar$ along which was then moved). To summarize
intuitively, given a sample designated to be the doubly-active
sample, as much movement as possible is made (along a so-called
``DRD'' direction computed for that sample) while decreasing the
objective function and keeping that sample doubly-active, and after
that movement, if the movement was interrupted by the presence of a
``blocking constraint'', at the point of interruption a switch in
movement direction takes place to the DRD of the new sample; i.e. a
switch from a single sample being margin-setter to a different
single sample being margin-setter. Our Lemmas and Theorem below show
that this approach is synonymous with fulfilling $R1$ and $R2$ that
were discussed in Sec. \ref{sec:descent:intro}.

The ``doubly-active'' property of a sample is represented and
notated as follows. In our ICQP problem (\ref{eqn:qp-linear}),
wherein each sample $\bxn$ is represented by a pair of companion
constraints $y_n ({a\bw}^{\rm T}\bxn + b) \geq 1-\xi_n$ and $\xi_n
\geq 0$, each pair contributes two rows to $\bA$ and there are $N$
samples (or pairs), and thus $\bA$ has a total of $2N$ rows.
Notationwise, the row arrangement we consider for the $2N$-row $\bA$
is that the top $N$-row half and the bottom $N$-row half are
respectively formed by the first constraint type ($y ({a\bw}^{\rm
T}\bx + b) \geq 1-\xi$) and the second constraint type ($\xi \geq
0$). As we elaborate shortly, similarly the row arrangement we
consider for the $\bAW$ matrix is a block arrangement with three
blocks (instead of two seen above for $\bA$) which, from top to
bottom, correspond to the three categories that samples fall into
according to whether $W$ (at an iteration of the active-set
algorithm) contains 1) only the first-type constraint for the sample
2) both constraint types for the sample 3) only the second-type
constraint for the sample. The second category here is designating
the doubly-active property of a sample. Shortly we elaborate on the
notation.

Restriction 1 on the size of set $W$ raises the nontrivial question
about how the particular constraints (no greater than $N+2$) for the
initial set $W$ should be selected at algorithm initialization among
all $2N$ constraints. During this initialization, recall that
essentially a classifier, specified (defined) by the following two
pieces of information, is input into the algorithm;{\it i)} the
particular set $P$ of training samples that were assigned positive
Lagrange multipliers by some classifier generator and {\it ii)} the
values of those multipliers. Consider, first, that this classifier
generator may or may not explicitly provide identification
information that identifies a particular sample $\bxn$ within that
particular set $P$ as being the ``margin-setter'' sample, that is,
the sample that the initial iteration of the active-set algorithm
would utilize as being the (initial) doubly-active sample. Here is
one instance where this identification is not provided by the
classifier generator; this generator, which may be a QP solver such
as LIBSVM, may assign, as we have experienced when using LIBSVM
(albeit with scalar training data), the value $C$ to {\it all}
multipliers within that set; the inconvenience that this scenario
brings is that the generator, by assigning to every sample in $P$
the same multiplier, is not indicating which samples in the set $P$
are the margin violators, unlike the alternative scenario wherein
margin violators become identified by the generator via the means of
setting to $C$ the multipliers for {\it only some of the samples in
$P$} (with the multipliers of remaining $P$ samples assigned a value
less than $C$, so as to identify those as ``the sample(s) at the
margin'' as opposed to margin violators).\footnote{Before
continuing, the reader could recall that this fact about all
positive multipliers being upper-bounded by $C$ is a characteristic
of the soft-margin SVM; see e.g. \cite{Burges_SVMtut}.} Consider,
second, that in some cases, when $f(\bxn)$ is computed under the
provided information {\it i} and {\it ii} above, it may
unfortunately be that the discriminant function value $y_n f(\bxn)$
does not compute precisely to $1$ for any $\bxn$ within set $P$,
such as seen in our experience with LIBSVM, and thus trying to
identify which samples in $P$ have their $y_n f(\bxn)$ equal to $1$
is not a reliable means, either (for determining which samples are
at the margin (or are doubly-active) and which other samples
aren't). Thus, extra measures may need to be taken to make that
determination. In particular, in the event that one finds out that
{\it i)} the generator that is generating and providing a classifier
as input into our active-set algorithm has happened to set all
positive multipliers to $C$ and {\it ii)} the discriminant $y_n
f(\bxn)$ is not computing to $1$ for any of those samples (with
those positive multipliers), a normalization measure can be taken
whereby one can utilize a scaling variable to scale to $1$ the
particular $y_n f(\bxn)$ that is both 1) the largest among the
particular $\bxn$ that have the positive multipliers and 2) positive
(to ensure that that particular $\bxn$ is a correctly classified
sample.) To summarize this paragraph, it is possible, by taking
measures, to provide to the active-set algorithm the designation of
what the algorithm's initial doubly-active sample is or could be,
even in the event there may seem to be potential numerical
obstacles; once this initial designation is made, the algorithm can
proceed as described above i.e. by essentially largely
mode-switching between {\it i)} moving along the DRD of a current
doubly-active sample and {\it ii)} when becomes necessary (i.e. when
a ``blocking constraint'' is encountered along the movement path),
switching to a new doubly-active sample so as to then move along its
DRD. We show below that this approach is computationally low-cost.
Specifically, the computational complexity at an ICQP iteration is
essentially the complexity of computing {\it the single basis
vector} for the null space $N(\bAW)$ of a highly sparse $N+1 \times
N+2$ matrix $\bAW$.\footnote{The reader can easily conclude, from
linear algebra, that the null-space of an $N+1 \times N+2$
full-row-rank matrix is one-dimensional and a subspace of
$\mathbb{R}^{N+2}$ and thus it has a single basis vector that has
$N+2$ coordinates.}

Shortly, in Lemma 3 and Lemma 4, respectively, we show that 1)
fulfilling {\it R1} requires that our $W$ {\it not contain both
constraints of a sample for more than two samples} and that 2)
fulfilling {\it R2} requires {\it every sample} to be represented in
$W$ (i.e. $W$ contains at least one of two constraints of every
sample, whereas, by contrast, A16.1 modifies $W$ freely without this
requirement since it addresses a more general case). Regarding how
{\it R1} and {\it R2} can be fulfilled, our theorem shows shortly
that $W$ would need to contain exactly either $N+1$ or $N+2$
constraints (from among the $2N$ constraints of the ICQP) that are
i) linearly independent and ii) include a constraint for each of the
$N$ samples. This points out that via specialization a more specific
Working Set strategy has emerged for QP1 from A16.1's; i.e. A16.1
allows, by contrast, $\bAW$ to have {\it fewer than} $N+1$ rows so
long as they are linearly independent. To summarize, as part of
specializing A16.1 to our particular QP problem QP1, our theorem is
extending A16.1's $W$ strategy, making it become more specific.

In preparation for the lemmas, we now elaborate on the notation
introduced above. We consider the index sets $W_1$ and $W_2$ that
respectively specify which rows of $\bA$'s top $N$-row half (for the
constraints of the form $y ({a\bw}^{\rm T}\bx + b) \geq 1-\xi$) and
bottom $N$-row half (for the constraints of the form $\xi \geq 0$)
form $\bAW$. Note that $W_1$ and $W_2$ contain relative (not
absolute) row indexes for $\bA$\footnote{The set of absolute indexes
of the $\bA$ rows that form $\bAW$ is given by $W\equiv W_1 \cup
(W_2 + N)$ where the plus sign denotes elementwise addition.}, which
are also sample indexes that specify, respectively, the samples that
have their first constraint in $W$ and the samples that have their
second constraint in $W$. A sample (sample index) whose both
constraints are in $W$ is in both $W_1$ and $W_2$, i.e. in
$W_{12}\equiv W_1 \cap W_2$; a sample being in $W_{12}$ means it is
a doubly-active sample (aka ``margin-setter''), whereas note that
both constraints of a sample may be active without the sample being
in $W_{12}$.\footnote{This is because, by definition of the Working
Set $W$, a constraint may be active without being in $W$.} Denoting
$W_{22}\equiv W_2 \setminus W_{12}$, the abovementioned three-block
structure for $\bAW$ is $\AWblocks$, where
$\bAWoneone\equiv[\bVoneone_{m_{11} \times 2} \hspace{0.05in}
\bIoneone_{m_{11} \times N}]$ (where $m_{11}\equiv card(W_{11})$),
$\bB\equiv \left(
  \begin{array}{cc}
    \bAWonetwo \\
    \bzero_{m_{12} \times 2} \hspace{0.3in} \bIonetwo_{m_{12} \times N} \\
  \end{array}
\right)$
(where $m_{12}\equiv card(W_{12})$), and
$\bAWtwotwo\equiv[\bzero_{m_{22} \times 2} \hspace{0.05in}
\bItwotwo_{m_{22} \times N}]$ (where $m_{22}\equiv card(W_{22})$).

{\it \textbf{Lemma 1:}} Collectively, rows of {\it i)} $\bAWoneone$,
{\it ii)} $\bAWtwotwo$, and {\it iii)} the bottom half of $\bB$
(i.e. $[\bzero \hspace{0.1in} \bIonetwo]$, are linearly independent.

{\it \textbf{Proof:}} Among the final $N$ columns of these rows, the
column at which $1$ appears (with the other $N-1$ columns being $0$)
differs from row to row. Q.E.D.

{\it \textbf{Lemma 2:}} If an $m\times n$ matrix $\bM$ is widened by
placing into it $k$ zero-columns (at column indexes
$i\in\{1,\ldots,n+k\}$ specified by a set $I$), a new complete set
of null-space basis vectors can be constructed from the old {\it
without computation} in two basic steps: 1) Grow each old $n \times
1$ basis vector to $(n+k)\times 1$, with 0s placed at coordinates
$i\in I$. 2) Into the basis vector set, additionally put $\bui$ for
each $i\in I$ (where $\bui$ is the special unit vector with $0$s
except $1$ at {\it i}-th coordinate). Consequently, in the new
$\bZ$, each row $j\notin I$ is the corresponding row of the old
$\bZ$, augmented {\it with only 0s}.

{\it \textbf{Proof:}} Right-multiplying a matrix by a (column)
vector produces the weighted sum of the matrix columns, with weights
being the vector elements. Thus: 1) The fact that the outcome of
right-multiplying $\bM$ by one of its null-space basis vectors is a
zero-vector remains unchanged in the event $\bM$ is widened and that
basis vector augmented with $0$s, because these $0$s do not
contribute to the weighted sum (the weighting occurs between the
complete set of original weights and original (pre-widening) columns
of $\bM$, with the result thus being the zero-vector). 2)
Right-multiplying the widened $\bM$ matrix by $\bui$ simply produces
the {\it i}-th column of that matrix, which (by definition) is a
zero-vector. Q.E.D.

{\it \textbf{Lemma 3:}} Fulfilling {\it R1} requires $W_{12}$ to
contain at most two samples (indexes).

{\it \textbf{Proof:}} $\bAWonetwo$, which is $[\bVonetwo_{m_{12}
\times 2} \hspace{0.05in} \bIonetwo_{m_{12} \times N}]$, is the top
half of $\bB$. Subtracting the bottom half of $\bB$ from the top
half of $\bB$ reduces $\bB$ to $\Bblockspostsub$ whose top half
$[\bVonetwo \hspace{0.05in} \bzero]$ can be row-rearranged as
$\left(
  \begin{array}{cc}
    \bVonetwoind & \bzero \\
    \bVonetwodep & \bzero \\
  \end{array}
\right)$ where $\bVonetwodep$ and $\bVonetwoind$ denote the two
blocks composed of, respectively, linearly dependent and independent
rows of $\bVonetwo$. Since $\bVonetwo$ has two columns, the number
of rows of $\bVonetwoind$ is at most two. Because the last $N$
columns in the above top half $[\bVonetwoind \hspace{0.05in}
\bzero]$ are zero, each row in this top half is linearly independent
of the rows in set $J$, where $J$ denotes the set of rows in
$\bAWoneone$ and $\bAWtwotwo$, due to Lemma 1. Collectively, again
due to Lemma 1, the rows in set $J$ and the rows of the bottom half
$[\bzero \hspace{0.1in} \bIonetwo]$ of $\bB$ are linearly
independent because among the final $N$ columns of all of these rows
the column at which $1$ appears, with the other $N-1$ columns being
$0$, differs from row to row. Thus, the linearly dependent rows of
$\bAW$, if any, are the rows of the above bottom block
$[\bVonetwodep \hspace{0.05in} \bzero]$. Q.E.D.

{\it \textbf{Lemma 4:}} Fulfilling {\it R2} requires every sample
$n$ to be represented in $W$, i.e. $n \in W_1\cup W_2
\hspace{0.05in} \forall n$.

{\it \textbf{Proof:}} Proof by contradiction. Suppose $n\notin
W_1\cup W_2$ for some sample $n$. Then, column $n+2$ of $\bAW$ is a
zero-column. Since such columns have column index $> 2$, by Lemma 2
each of rows $n\leq 2$ of $\bZ$ is the corresponding row of
$\bZtilde$ augmented {\it with only 0s} (where $\bZtilde$ is the
null-space matrix that would result from first removing the
zero-columns of $\bAW$); i.e. $\bzrowone$, the first row of $\bZ$,
contains at least one $0$. Thus, $\bzrowone^{{\rm T}} \bzrowone$,
which is real and symmetric, has 0 as an eigenvalue. Since
$\bZ^{{\rm T}} \bG \bZ$ is $||w^{-\Mcal}||^2 \bzrowone^{{\rm T}}
\bzrowone$, $\bZ^{{\rm T}} \bG \bZ$ too is real and symmetric and
has 0 as an eigenvalue. Q.E.D.

{\it \textbf{Theorem 1:}} At each ICQP iteration $k$, fulfilling
{\it R1} and {\it R2} would require $W^k$ to contain exactly either
$N+1$ or $N+2$ constraints (from among $2N$ ICQP constraints) that
{\it i)} are linearly independent and {\it ii)} include a constraint
for each of the $N$ samples.\footnote{Recall from above that A16.1
\cite{Nocedal} allows, by contrast, $\bAW$ to have {\it fewer than}
$N+1$ rows, so long as they are linearly independent; i.e. our
Theorem has introduced a specialization of A16.1, extending its
Working Set strategy.} We refer to these conditions as {\it C1} and
{\it C2}.

{\it \textbf{Proof:}} By Lemma 3, fulfilling $R1$ requires $W_{12}$
to be (a) empty or contain either (b) one or (c) two samples (sample
indexes); i.e., $\bAWonetwo$ is required to respectively be (a)
empty or contain either (b) one or (c) two (linearly independent)
rows. Under these three options wherein $\bAW$ is full-row-rank with
at least $N$ (linearly independent) rows due to collectively Lemmas
1, 3, 4, $\bZ$ is, respectively, two-column ($\bAW$ has $N$ rows) or
one-column ($\bAW$ has $N+1$ rows) or empty ($\bAW$ has $N+2$ rows).
Option 1 does not fulfill {\it R2} because, when the first row
$\bzrowone$ of $\bZ$ is two-column (i.e. $\bzrowone\equiv[z_1
\hspace{0.05in} z_2]$), $\bzrowone^{{\rm T}} \bzrowone \equiv \left(
  \begin{array}{cc}
    z_1^2 & z_1 z_2 \\
    z_1 z_2 & z_2^2 \\
  \end{array}
\right)$ is not positive definite and thus neither is $\bZ^{{\rm T}}
\bG \bZ$. This requires $W_{12}$ (and thus both $W_1$ and $W_2$) to
be nonempty. Under option 2, $\bzrowone$ is a scalar $z$ and {\it
R2} is fulfilled because $\bZ^{{\rm T}} \bG \bZ$, which is
$(||w||^2)^{-\Mcal} z^2$, is a positive scalar (and thus positive
definite). Under option 3, wherein {\it R2} does not apply (since
$\bZ$ is empty), the KKT matrix is nonsingular because $\bAW$ is.
Thus the option needs to be option 2 or 3; i.e. $\bZ$ is one-column
if not empty. Hence the two conditions stated in the Theorem. Q.E.D.

While the optimization parameter set is $\{a,b,\xi_1,\ldots,\xi_N\}$
(i.e. includes slackness parameters $\xi_n$ apart from the parameter
pair ($a$,$b$)), at any iteration of the active-set algorithm the
($a$,$b$) part of the parameter set
$\{a$,$b$,$\xi_1$,$\ldots$,$\xi_N\}$ sufficiently specifies a found
solution (where ``found solution'' means a feasible point that is
either an interim iteration-specific solution or the found final
solution for the algorithm), so long as it is ensured that the set
$W_{12}$ is not empty at that solution i.e. so long as there is a
sample $\bxn$ ensured to be the single doubly-active sample
discussed above. Utilizing Theorem 1, an active-set algorithm can
ensure this. This may involve a (within-iteration) scaling of the
$(a,b)$ (i.e. $\bqonetwo$) part of $\bq$ so as to make equal to $1$
the discriminant function value $y_n f(\bxn)$ for that sample $\bxn$
(since $\bqonetwo$ is the only part of $\bq$ that determines that
$y_n f(\bxn)$ value), since $y_n f(\bxn)$ being equal to $1$ does
ensure the sample is doubly-active. Upon such scaling via
utilization of the $\bqonetwo$ part of $\bq$, in order to
accordingly adjust the remaining $\bq$ coordinates the new
slacknesses $\xi_l=\max(0,1-g_l) \hspace{0.05in} \forall l$ can be
calculated.

Now we discuss our specialization of A16.1 so as to fulfill {\it C1}
and {\it C2}, as well as show the constraint selection discussed in
Theorem 1 is straightforward and computationally low-cost; in
particular, we show we reduce the $\bpk$ computation to only
$\approx$ $N$ additions and $3$ multiplications, given $\bZk$. At
initial iteration ($k=0$), using the boundary (i.e. $a=1$,
$b=w_{0}^{-\Mcal}$) information being input to the algorithm (aka,
as we discussed earlier, the information identifying both a
particular set $P$ of samples as having positive Lagrange
multipliers and the values of those multipliers), one of the
correctly classified samples $\bxn \in P$ can be made and designated
doubly-active (with scaling performed if necessary) and placed into
$W_{12}$; accordingly, the other $N-1$ samples, upon computing
slackness for them, can be placed into either $W_1$ or $W_2$ based
on individual slackness value. Such $W^0$ has $N+1$ samples and
fulfills {\it C1} and {\it C2}. Subsequently, for $k>0$, to fulfill
{\it C1} and {\it C2}, our specialization of $W$ modifications is as
follows. We only need to compute $\bpkstar$ when $W$ contains $N+1$
constraints\footnote{Because in the $N+2$ case $\bZ$ (and thus
$\bpkstar$) is $\bzero$; cf. proof of Theorem 1.}, in which case
$\bZk$ is one-column ($\bZk\equiv[z_1 \ldotswithhspace
z_{N+2}]^{{\rm T}})$; i.e. the null space $N(\bAW)$ of $\bAW$ is a
one-dimensional subspace of $\mathbb{R}^{N+2}$. To compute
$\bpkstar$, we employ the ``null-space method'' used for solving
ECQP problems \cite{Nocedal}; the approach is based on taking an
input $\bpkin \in N(\bAW)$ (that fulfills $\bAW\bpkin=\bzero$ of
(\ref{eqn:ecqp}); e.g. due to $\bZk$ being one-column, $\bpkin$ can
be set to a multiple of $\bZk$, e.g. $\bZk$ itself) and computing a
displacement vector $\bn$, with $\bpkstar=\bpkin+\bn$. Since
$\bpkstar$ must fulfill $\bAW\bpkstar=\bzero$ (\ref{eqn:ecqp}) (i.e.
$\bpkstar$ must be in $N(\bAW)$), we can see that $\bn$ can too be
(i.e. like $\bpkin$) in $N(\bAW)$ and thus $\bn$ can be expressed as
$\bZk\bnZ$ (and may be $\bzero$ or nonzero) for some column-vector
$\bnZ$. Finding $\bnZ$ takes almost no computation because
$(\bZk^{{\rm T}} \bG \bZk)\bnZ = -\bZk^{{\rm T}}\brk$ \cite{Nocedal}
where
both $\bZk^{{\rm T}} \bG \bZk$\footnote{$\bZk^{{\rm T}} \bG \bZk$
may seem computationally costly but it is not, because, due to
$\bZk$ being one-column, $\bZk^{{\rm T}} \bG \bZk$ is given by the
scalar $(||\bw||^2)^{-\Mcal} z_1^2$ which is not computationally
costly.} (which is $(||\bw||^2)^{-\Mcal} z_1^2$) and $\bZk^{{\rm
T}}\brk$ are known scalars (and thus so is $\bnZ$, i.e. $\bnZ=n_z$);
here, $\brk\equiv\bhk+\bG\bpkin=[(||\bw||^2)^{-\Mcal}
(\bqk_{1}+\bpkin_{1})$\footnote{Subscript $i$ as used here on $\bqk$
and $\bpkin$ for $i=1$ denotes {\it i}-th
coordinate.}$\hspace{0.05in} 0 \hspace{0.05in} C\bone_{1\times
N}]^{{\rm T}}$. Thus when $\bpkin$ is set to $\bZk$ itself (as
mentioned above) the $\bpkstar$ can be found by simply multiplying
$\bZk$ by the scalar
$-(\frac{\bqk_{1}}{z_1}+\frac{C}{(||\bw||^2)^{-\Mcal}
z_1^2}\sum\limits_{i=3}^{N+2}z_i)$. We have thus reduced, as
mentioned above, the $\bpkstar$ computation to only $\approx$  $N$
additions and $3$ multiplications, given $\bZk$. We have thus
reduced, as mentioned above, the complexity of the $\bpkstar$
computation to the complexity of finding the single basis vector
$\bZk$ for iteration $k$. After finding $\bpkstar$, one can compute
$\delta_k$, determine $j\equiv \mbox{arg}\min\limits_{i\in
B}(1,R_i)$ and associated sample $n_j$, upon which two cases need to
be considered. In the first case, defined as $\delta_k$ being $1$,
there is no blocking constraint (i.e. one can move from $\bqk$ to
$\bqk+1\bpkstar$ without interruption); accordingly the sample
currently designated ``doubly-active'' can maintain that designation
and $W$ can remain the same. In the second case, defined as
$0<\delta_k<1$, there is a blocking constraint, which is a
constraint associated with $n_j$, and thus $n_j$ can become the
newly designated doubly-active sample; i.e. $W$ can be modified by
placing $j$ into $W$ to replace in $W$ a constraint of the previous
doubly-active sample. At any setting, or designation, of a sample as
``doubly-active'' sample, slackness may need to be computed for the
remaining $N-1$ samples to ensure that all slackness values are
consistent with the contents of $W$. \vspace{-0.1in}
\bibliography{refs}

\begin{thebibliography}{99}
\bibitem{Aksu_TNN}
Y. Aksu, D. J. Miller, G. Kesidis, Q. X. Yang, ``Margin-maximizing
feature elimination methods for linear and nonlinear kernel-based
discriminant functions'', {\it IEEE Trans. Neural Netw.}, vol. 25,
no.10, pp.701-717, 2010.
\bibitem{Aksu_PLOS}
Y. Aksu, D. J. Miller, G. Kesidis, D. C. Bigler, Q. X. Yang, ``An
MRI-derived definition of MCI-to-AD conversion for long-term,
automatic prognosis of MCI patients'', {\it PLoS ONE}, 6(10):e25074,
2011.
\bibitem{Burges_SVMtut}
C. J. C. Burges, ``A tutorial on support vector machines for pattern
recognition,'' {\it Data Mining Knowledge Disc.} 2 (2) pp.121-167,
1998.
\bibitem{Libsvm}
C. Chang, C. Lin, ``LIBSVM: a library for support vector machines,''
software available at http://www.csie.ntu.edu.tw/$\sim$cjlin/libsvm,
2001.
\bibitem{Chung-RadiusMarginBound}
K-M. Chung, W-C. Kao, C-L. Sun, L-L. Wang, C-J. Lin, ``Radius Margin
Bounds for Support Vector Machines with the RBF Kernel,'' full
version (Manuscript Number: 2643); ICONIP'02 (Vol.2) and Neural
Computation (Vol. 15, No. 11, Pages 2643-2681, Nov. 2003).
\bibitem{Cover_ITEC}
T. M. Cover, ``Geometrical and statistical properties of systems of
linear inequalities with applications in pattern recognition,'' {\it
IEEE Trans. on Electronic Computers}, EC-14(3), pp.326--334,
1965.
\bibitem{Dasgupta_FE}
S. Dasgupta, Y. Goldberg, M. R. Kosorok, ``Feature Elimination in
Empirical Risk Minimization and Support Vector Machines,'' {\it The
University of North Carolina at Chapel Hill Department of Statistics
Technical Report Series}, Paper 37, Year 2013.
\bibitem{Duda}
R. Duda, P. Hart, G. Stork, {\em Pattern Classification}, Wiley \&
Sons, 2001.
\bibitem{Hastie_book}
T. Hastie, R. Tibshirani, J. Friedman, {\em The Elements of
Statistical Learning}, Second Edition, Springer, 2011.
\bibitem{Nocedal}
J. Nocedal, and S. J. Wright, {\em Numerical Optimization},
Springer, 1999.
\bibitem{Su_2002}
Y. Su, T. M. Murali, V. Pavlovic, S. Kasif, ``Training support
vector machines in 1D'', Dec. 8, 2002; tech. report at author
website.
\bibitem{Vapnik_1995}
V. Vapnik, {\em The Nature of Statistical Learning Theory}, Springer
Verlag, New York, 1995.
\bibitem{Vapnik_1998}
V. Vapnik, {\em Statistical Learning Theory}, John Wiley \& Sons,
1998.
\bibitem{Vapnik-Chapelle}
V. Vapnik, O. Chapelle, ``Bounds on error expectation for support vector machines,'' {\it Neural Computation}, 12(9):2013-2036.
\bibitem{Weston_NIPS}
J. Weston, S. Mukherjee, O. Chapelle, M. Pontil, T. Poggio, V.
Vapnik, ``Feature selection for SVMs,'' {\it NIPS 13}, MIT Press,
2001.
\bibitem{Weston_thesis}
J. Weston,  ``Extensions to the Support Vector Method,'' Ph.D.
thesis, University of London, October 1999.
\bibitem{Zhang_radius}
T. Zhang, ``A leave-one-out cross validation bound for kernel
methods with applications in learning,'' In {\it Proceedings of the
14th Annual Conference on Computational Learning Theory}, vol. 2111,
pages 427-443, Berlin, NY, Springer.
\end{thebibliography}

\vspace{-0.1in}
\end{document}